\pdfoutput=1

\documentclass[11pt]{article}

\usepackage[final]{acl}

\usepackage{times}
\usepackage{latexsym}

\usepackage[T1]{fontenc}

\usepackage[utf8]{inputenc}

\usepackage{microtype}

\usepackage{inconsolata}

\usepackage{multirow} 
\usepackage{booktabs} 
\usepackage{bbding}
\usepackage{graphicx}
\usepackage{amssymb}
\usepackage{colortbl}
\usepackage{amsmath} 
\usepackage[normalem]{ulem}
\usepackage{inconsolata}
\usepackage{arydshln}
\useunder{\uline}{\ul}{}

\title{Detection-Correction Structure via General Language Model for Grammatical Error Correction}

\author{Wei Li, Houfeng Wang \\
  National Key Laboratory for Multimedia Information Processing, Peking University \\
  weili22@stu.pku.edu.cn, wanghf@pku.edu.cn}
  
\begin{document}
\maketitle
\begin{abstract}
Grammatical error correction (GEC) is a task dedicated to rectifying texts with minimal edits, which can be decoupled into two components: detection and correction. However, previous works have predominantly focused on direct correction, with no prior efforts to integrate both into a single model. Moreover, the exploration of the detection-correction paradigm by large language models (LLMs) remains underdeveloped. This paper introduces an integrated detection-correction structure, named DeCoGLM, based on the General Language Model (GLM). The detection phase employs a fault-tolerant detection template, while the correction phase leverages autoregressive mask infilling for localized error correction. Through the strategic organization of input tokens and modification of attention masks, we facilitate multi-task learning within a single model. Our model demonstrates competitive performance against the state-of-the-art models on English and Chinese GEC datasets. Further experiments present the effectiveness of the detection-correction structure in LLMs, suggesting a promising direction for GEC.
\end{abstract}


\section{Introduction}

Grammatical error correction (GEC) is a task focused on automatically rectifying grammatical errors in human-written text \citep{wang2021comprehensive_survey_gec}. GEC models are applied in language learning \citep{katinskaia2021assessing_in_learning, caines2023teaching_application_gec, kaneko2022interpretability_of_gec}, enhancing automatic speech recognition \citep{liao2023improving_asr_by_gec}, and aiding in text data labeling \citep{sun2023htec_text_data_by_gec}. The two primary approaches in GEC are Sequence-to-Sequence (Seq2Seq) and Sequence-to-Edit (Seq2Edit). Without detection, Seq2Seq treats GEC as the direct generation for correct text, providing high flexibility \citep{junczys2018first_transformer_for_gec, ge-etal-2018-fluency_gec}. On the other hand, Seq2Edit views GEC as a sequence labeling task for edit labels, showcasing high precision by controlled edits \citep{awasthi2019parallel_edit_seq2edit_gec, stahlberg2020seq2edits_gec, omelianchuk2020gector_gec}. The advent of large language models (LLMs) has further expanded Seq2Seq model capabilities \citep{ouyang2022GPT3_5_RLHF, zeng2022glm_130b}. Despite their unprecedented performance in various tasks \citep{chang2023survey_eval_llm}, LLMs underperform than low-parameter models in GEC due to the over-correction phenomenon \citep{qu2023llm-chinesegec, coyne2023gpt-gec-en}.

\begin{figure}[t]
    \centering
    \includegraphics[width=0.9\columnwidth]{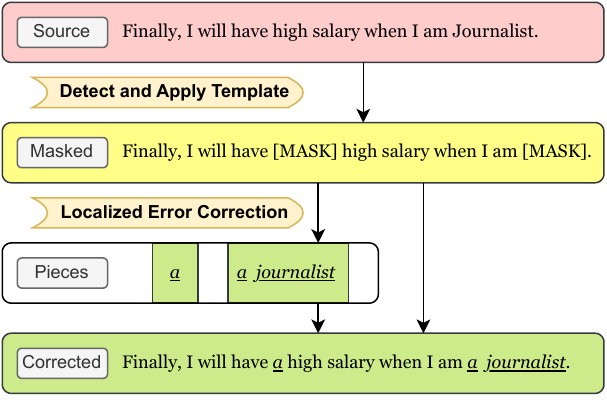}
    \caption{Detection and correction process of DeCoGLM. \textbf{Detection} and \textbf{Correction} are incorporated in one General Language Model (GLM). }
    \label{fig:intro}
\end{figure}

While the detection-correction structure can harness the strengths of both Seq2Seq and Seq2Edit, most existing works merely utilize detection as additional input for Seq2Seq models \citep{yuan-etal-2021-two_systems_detection_correction_gec, li2022seq2action, li-etal-2023-templategec}. Moreover, all previous detection-correction systems comprise separate models \cite{chen-etal-2020-seq2span_gec}. In contrast, we introduce a novel GEC model, named DeCoGLM, based on the General Language Model (GLM) \citep{du2021glm}. This model employs an integrated detection-correction structure to detect errors and generate localized corrections. As depicted in Figure \ref{fig:intro}, the error detection phase employs a template rule to construct masked text based on detection results. During the correction phase, the model leverages the autoregressive mask infilling capability of the GLM to generate correct text pieces for erroneous parts, thereby saving inference time. 
To incorporate both detection and correction within a single model, we devise a multi-task learning approach, organizing input text with attention mask adjustments. Results on English and Chinese GEC benchmarks demonstrate that our proposed model surpasses previous detection-correction models and is comparable to state-of-the-art (SOTA) models. To further explore the potential of applying the detection-correction structure to LLMs, the detection and correction phases are separated, termed DeGLM and CoGLM respectively. 
Our proposed single system, comprising a small 
detection model and an LLM corrector, outperforms other Seq2Seq LLMs. In summary, our primary contributions are:

\begin{itemize}
    \item A novel GEC model, DeCoGLM, which incorporates a detection-correction structure based on the GLM.
    \item The design of a multi-task training method that integrates detection and correction within a single model.
    \item The exploration of using LLMs for GEC, which involves deploying large error correction models with the support of 
    small detection models.
\end{itemize}

\section{Related Work}

\subsection{Sequence-to-Sequence GEC}

Seq2Seq models \citep{2019bart, 2020t5} have demonstrated high performance in GEC \citep{junczys2018first_transformer_for_gec, choe-etal-2019-transferlearning_gec, zhao-etal-2019-seq2seq-copy-augmented, katsumata-komachi-2020-bart-basseline-gec}. Techniques such as data synthesis \citep{stahlberg-kumar-2021-synthetic-tag-corrupted, grundkiewicz2019synthetic_data_gec}, training schedule \citep{lichtarge2020dataweightedtraining_gec, bout-etal-2023-training-schedule}, and decode reranking methods \citep{kaneko2019transformer_reranking_for_gec, zhang-etal-2023-transformer-reranker, zhou-etal-2023-improving-seq2seq-decoding-interventions} have been incorporated into previous Seq2Seq GEC models. SOTA model architectures typically supplement Seq2Seq models with additional information \citep{li-etal-2023-templategec, zhang-etal-2022-syngec, fang-etal-2023-multimodal-gec}. However, a significant drawback of Seq2Seq GEC models is the inference cost, as these models generate tokens sequentially and waste time copying source tokens \citep{sun-etal-2021-instantaneous-gec}.

As the latest Seq2Seq models, LLMs have emerged as a new paradigm for natural language processing (NLP) tasks following the introduction of GPT-3 and ChatGPT \citep{brown2020gpt3}. Nevertheless, recent studies have shown that LLMs underperform current SOTA models on both English and Chinese GEC benchmarks \citep{coyne2023gpt-gec-en, loem2023gpt3-gec-en, qu2023llm-chinesegec, li2023gpt-chinesegec}. Existing datasets and evaluation methods \citep{bryant-etal-2017-errant} favor minimum edits as the rule for correction. However, GPT models often produce over-corrected sentences with unnecessary edits \citep{fang2023chatgpt-gec-en, coyne2023gpt-gec-en}. In contrast to the Seq2Seq GEC methods that directly perform overall generation, our work only focuses on localized error correction, which not only saves inference time but also mitigates the over-correction phenomena in LLMs.

\subsection{Sequence-to-Edit GEC}

Seq2Edit methods generate edit operations for ungrammatical sentences \citep{stahlberg2020seq2edits_gec}. For instance, LaserTagger \citep{malmi-etal-2019-lasertagger} predicts token-level edit operations, which has been adopted in subsequent methods like PIE and GECToR \citep{awasthi2019parallel_edit_seq2edit_gec, omelianchuk2020gector_gec}. As a representative model, GECToR predicts four classes of edits and grammatical transformations, achieving high-precision results. \citet{lai-etal-2022-type-driven-multi-turn} further enhances it by addressing its deficiencies in multi-round correction. However, Seq2Edit methods necessitate intricate designs for edits, which are not language-agnostic. In contrast, our proposed model retains a limited set of language-agnostic edit operations and can flexibly conduct edits by autoregressive generation.

\begin{figure*}[ht]
    \centering
    \includegraphics[width=0.9\textwidth]{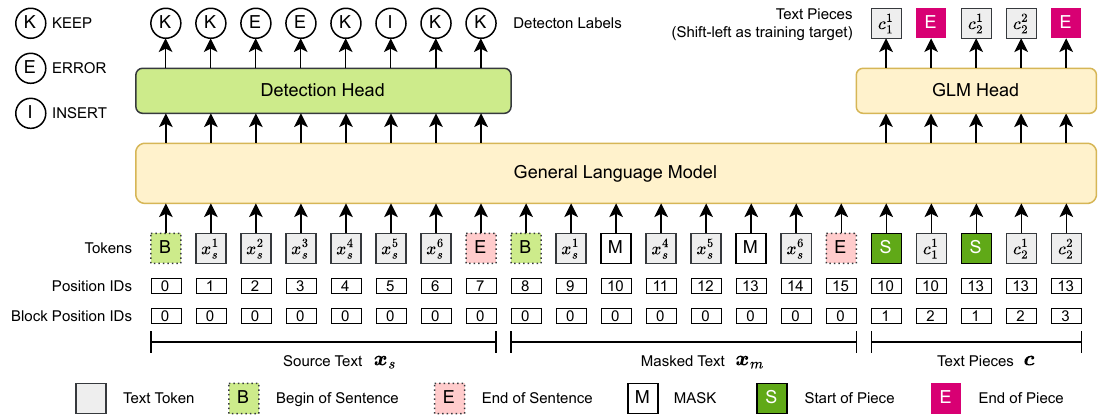}
    \caption{The proposed \textbf{detection-correction structure} based on GLM. The example shown above has a source text $\boldsymbol x_s = x_s^1 x_s^2 x_s^3 x_s^4 x_s^5 x_s^6$ and the target text is $\boldsymbol y = x_s^1 c_1^1 x_s^4 x_s^5 c_2^1 c_2^2 x_s^6$. Consistent with GLM, the position IDs and block position IDs are utilized for marking the original positions of text pieces and the inner order of tokens.}
    \label{fig:model}
\end{figure*}

\subsection{Detection-Correction GEC}

The GEC task can be divided into two processes: detection and correction \citep{rei2016compositional_detection_gec, bell2019context_detection_gec}. Prior research incorporates detection results as supplementary information for Seq2Seq correction models \citep{kaneko-etal-2020-bert-fused-ged, yuan2021multiclass_detection_correction, li-etal-2023-templategec}. The methods proposed by \citet{mallinson-etal-2020-felix-text-editing} and \citet{yakovlev-etal-2023-gec-depend} employ the Masked Language Model (MLM) \citep{2018bert} to obtain corrections, which are constrained by mask number. \citet{chen-etal-2020-seq2span_gec} introduces error span detection and correction to address the GEC problem, which allows for flexible corrections while maximizing time efficiency. Building on this, we further integrate the detection and correction tasks into a single GLM model, enabling mutual benefits between the two tasks, which is not achieved by previous works.

\section{Methods}

Our proposed model leverages the design of the GLM. Given a sentence with MASK tokens, GLM utilizes autoregressive blank infilling \citep{du2021glm} to generate a corresponding segment for each MASK position. These segments are termed as \textbf{text pieces}. This section describes how GLM is utilized to integrate detection and correction into a single model, as depicted in Figure \ref{fig:model}. Additionally, the design of multi-task training is also outlined here.

\subsection{Error Detection}
\label{sec:error-detection}
Drawing from the four edit classes by \citet{omelianchuk2020gector_gec}, we utilize token-level detection labels that do not include any specific word or grammar. Given that the mask-infilling process can generate empty text pieces, the REPLACE and DELETE operations are consolidated into the ERROR label. Consequently, the detection labels comprise KEEP ($K$), ERROR ($E$), and INSERT ($I$).
Given the tokens of \textbf{source text} as: 
\begin{equation}
\boldsymbol x_s = x_s^1 x_s^2 \dots x_s^n
\label{eq:source-text}
\end{equation}
, the objective of error detection is to predict \textbf{detection labels} derived by the alignment between the source text and the target text (correct text): 
\begin{equation}
\boldsymbol d = d_1 d_2 \dots d_n, d_i\in L = \left \{ K, E, I \right \}
\label{eq:detection-labels}
\end{equation}

\noindent \textbf{Detection Model} \quad The proposed model begins by extracting the representations of the source text tokens by $\mathrm{GLM}$ as Equation \ref{eq:source-text-representations}. The final detection label predictions are generated through a detection head, implemented by a feed-forward network $\mathrm{FN}$ and softmax function, as shown in Equation \ref{eq-detection-predictions}:
\begin{equation}
\boldsymbol h_s = h_s^1 h_s^2 \dots h_s^n = \mathrm{GLM}\left ( \boldsymbol x_s \right ) 
\label{eq:source-text-representations}
\end{equation}
\begin{equation}
p\left ( \hat{d_i} = l | \boldsymbol x_s \right ) = \mathrm{Softmax}(\mathrm{FN}\left ( h_s^i \right )), l\in L
\label{eq-detection-predictions}
\end{equation}

\noindent \textbf{Fault-tolerant Template} \quad The source text $\boldsymbol x_{s}$ is transformed into masked text $\boldsymbol x_{m}$ based on the detection labels using the following template rules. Each continuous interval containing only ERROR labels is replaced with a MASK token. For each position of INSERT, a MASK token is inserted. The form of \textbf{masked text} is shown in Equation \ref{eq:masked-text}:
\begin{equation}
\boldsymbol x_m = \boldsymbol x_{s_1} m_1 \boldsymbol x_{s_2} m_2  \dots m_k \boldsymbol x_{s_{k+1}},
\label{eq:masked-text}
\end{equation}
where $m_i$ is the $i$-th MASK token introduced in $\boldsymbol x_{s}$, and $\boldsymbol x_{s_i} $ denotes the $i$-th correct subinterval of source text.
If all the labels are KEEPs, the source text is directly output as the corrected result. Despite potential inaccuracies in detections, our model can tolerate a certain degree of false positives. In the instance where the correct token is identified as ERROR or INSERT, the corrector can mitigate such errors by either restoring the original text piece or generating an empty text piece.

\noindent \textbf{Aggressive Detection} \quad Utilizing the fault-tolerant template enables more aggressive detection, emphasizing the recall of ERROR and INSERT. Focal Loss \citep{lin2017focalloss} is used as the loss function to tackle the issue of imbalanced classification because the majority of tokens correspond to KEEP labels. The training objective for error detection is given by Equation \ref{eq:detection-loss}: 
\begin{equation}
\ell_D=-\boldsymbol\alpha_D \left ( 1-p_\theta \left ( \boldsymbol d|\boldsymbol x_s \right )  \right )^\gamma \mathrm{log} \left ( p_\theta \left ( \boldsymbol d|\boldsymbol x_s \right ) \right ) 
\label{eq:detection-loss}
\end{equation}
where $\theta$ represents the model parameters and $\gamma$ is a hyper-parameter set to 2. $\boldsymbol\alpha_D$ denotes the corresponding weight factors for detection labels. To strengthen aggressive error detection, $\alpha_K$ for the KEEP category is set to 1, while $\alpha_{EI}=2$ is set for the ERROR and INSERT categories.

\subsection{Localized Error Correction}
\label{sec:local-error-correction}

In the training data, detection labels are derived from the alignment of sequences between the source text $\boldsymbol x_s$ and the target text $ \boldsymbol y $. The corresponding masked text $\boldsymbol x_m$ can be formulated in Equation \ref{eq:masked-text} with $\boldsymbol x_{s_i}$ representing the $i$-th aligned segment. For each unaligned position replaced with $m_i$, the correct text piece is denoted as $ \boldsymbol c_i $. Consequently, the target text can be represented as:
\begin{equation}
\boldsymbol y = \boldsymbol x_{s_1} \boldsymbol c_1 \boldsymbol x_{s_2} \boldsymbol c_2  \dots \boldsymbol c_k \boldsymbol x_{s_{k+1}}
\label{eq:target-text}
\end{equation}

Leveraging the GLM pretrained by autoregressive blank infilling task, we fine-tune the GLMs for localized error correction. The probability distribution prediction for the $j$-th token in the $i$-th text piece $ \boldsymbol c_i $ is given in Equation \ref{eq:glm-predictions}:
\begin{equation}
\small
\label{eq:glm-predictions}
\begin{split}
p\left ( \hat{c_{i,j}} = w | \boldsymbol x_s, \boldsymbol x_m, \boldsymbol c_{<i}, \boldsymbol c_i^{<j} \right ) = \\
\mathrm{GLMH} \left ( \boldsymbol x_s, \boldsymbol x_m, \boldsymbol c_{<i}, \boldsymbol c_i^{<j} \right ), w\in V
\end{split}
\end{equation}
where $\mathrm{GLMH}$ denotes the GLM model with its original token prediction head, $w$ is any token in the vocabulary, and $\boldsymbol c_i^{<j}$ refers to all tokens with index $<j$ in text piece $\boldsymbol c_i$.

\subsection{Multi-Task Organization}
\label{sec:multi-task-organization}

\noindent \textbf{Multi-task Learning.} \quad The cross-entropy loss function, shown in Equation \ref{eq:glm-loss}, is used as the training objective for error correction task:
\begin{equation}
\ell_C=- \sum_{i,j}\mathrm{log} \left ( p_\theta \left ( c_{i,j} | \boldsymbol x_s, \boldsymbol x_m, \boldsymbol c_{<i}, \boldsymbol c_i^{<j} \right ) \right ) 
\label{eq:glm-loss}
\small
\end{equation}

For multi-task learning, we utilize a weighted loss function to enable the model to concurrently acquire error detection and correction capabilities. The training objective for this DeCoGLM model is to minimize the loss function given by:
\begin{equation}
\ell = \bar{\ell_C} + w_D \bar{\ell_D}
\label{eq:loss}
\end{equation}
where $\bar{\ell_C}$ and $\bar{\ell_D}$ are the token-level averages of $\ell_C$ and $\ell_D$ respectively. The detection loss weight $w_D$ is set to 10 to balance the scales of the two losses. For the impact of the loss weights on the model's performance, please refer to Section \ref{sec:weight-study}.

\begin{figure}[ht]
    \centering
    \includegraphics[width=0.9\columnwidth]{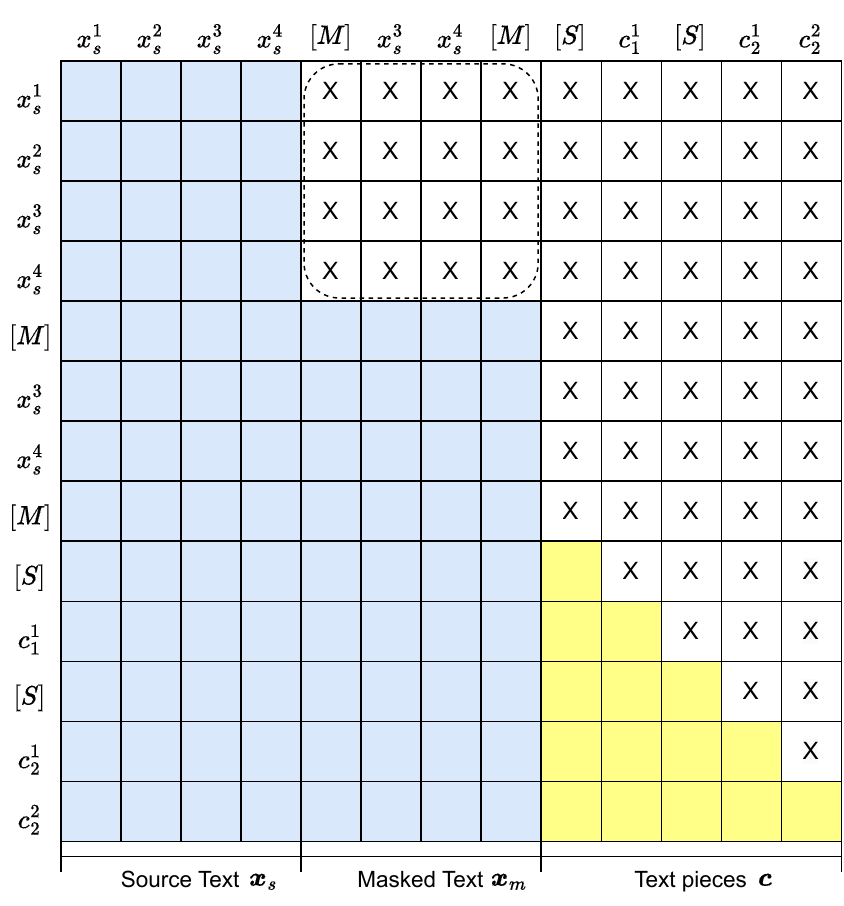}
    \caption{\textbf{Attention Mask Example.} The source text is $\boldsymbol x_s = x_s^1 x_s^2 x_s^3 x_s^4$, and the target text is $\boldsymbol y = c_1^1 x_s^3 x_s^4 c_2^1 c_2^2$. If the cell at row $i$ and column $j$ is colored, it indicates that the $i$ th token can pay attention to the $j$ th token. The region enclosed by dashed lines indicates the attention removed compared to the original GLM.
}
    \label{fig:attention}
\end{figure}

\noindent \textbf{Attention Mask} \quad To unify the two tasks into a single model, source text $\boldsymbol x_s$, masked text $\boldsymbol x_m$, and text pieces $\boldsymbol c$ are concurrently fed into the GLM model. The prediction of detection labels is conditioned on $\boldsymbol x_s$, while the autoregressive text prediction relies on $\boldsymbol x_s$, $\boldsymbol x_m$, and all previously generated text pieces. Therefore, the attention from $\boldsymbol x_m$ to $\boldsymbol x_s$ is eliminated to prevent detection from using $\boldsymbol x_m$ and $\boldsymbol c$, with other part adhering to the original GLM attention mask. This is depicted in Figure \ref{fig:attention}.

\noindent \textbf{Two Stage Supervised Fine-tuning} \quad Given that error detection is not infallible, the input during the correction phase may contain inaccuracies, with a distribution deviation from the training samples constructed with right detection labels. This issue is also observed in other detection-correction works \citep{chen-etal-2020-seq2span_gec, li-etal-2023-templategec}. To address this, we add a second supervised fine-tuning stage (SFT2), which employs a \textbf{detection-enhanced} approach: initially, all detection results on the training set are obtained using the model trained with data constructed by perfect detection (SFT1). Then, new training data is generated by augmenting the original labels with the fault detection results, leading to a secondary training of the SFT1 model. Examples of the two-stage training samples are provided in Table \ref{tab:data-examples} in the appendix.

\subsection{Separate Models}
The detection and correction phases can be implemented using two separate GLMs, named DeGLM and CoGLM respectively in this paper. Their training objectives are defined by Equation \ref{eq:detection-loss} and \ref{eq:glm-loss}, respectively. This decomposition facilitates the customization of distinct models for the detection and correction phases. However, in scenarios with limited computational resources, the Parameter-Efficient Fine-Tuning (PEFT) \citep{fu2023peft} is ineffective for DeCoGLM due to the significant disparities between the sequence labeling task of the error detection module and the mask-infilling pretraining task of GLM. To apply our approach to LLMs, we propose training a large version of CoGLM using the detection-enhanced method, similar to the second stage fine-tuning discussed in Section \ref{sec:multi-task-organization}.

\subsection{Detection Control}
\label{sec:detection-control}

During inference, the model needs to predict detection labels for the source text first, then transform it into masked text. Subsequently, both of them are input together for generating text pieces. This decoupling allows us to regulate the correction process using the probabilities of the three detection labels, thereby harnessing the model's potential to enhance benchmark performance. Three control modes are designed:

\noindent \textbf{KEEP Threshold} (\(\delta\)): Any prediction with KEEP probability exceeding \(\delta\) is directly set to KEEP.

\noindent \textbf{ERROR Lower Bound} (\(\phi_e\)): Any ERROR probability prediction falling below \(\phi_e\) is directly set to 0, thereby precluding the prediction of ERROR when \(p_e < \phi_e\).

\noindent \textbf{INSERT Lower Bound} (\(\phi_i\)): Any INSERT probability prediction below \(\phi_i\) is directly set to 0, precluding the prediction of INSERT when \(p_i < \phi_i\).

The three inference hyper-parameters can be determined using a greedy grid search based on the metrics on the validation set. We discuss them in Section \ref{sec:pre-rec-trade-off}.

\section{Experiments}

\begin{table*}
\centering
\resizebox{\textwidth}{!}{
\begin{tabular}{llcccccccccccc}
\hline
                                        &                                                        & \multicolumn{6}{c}{\textbf{English}}                                                                                                                                                          & \multicolumn{6}{c}{\textbf{Chinese}}                                                                                                                 \\ \cline{3-14} 
\textbf{}                               & \textbf{}                                              & \multicolumn{3}{c}{\textbf{CoNLL-14} \textit{test}}                                           & \multicolumn{3}{c}{\textbf{BEA-19} \textit{test}}                                             & \multicolumn{3}{c}{\textbf{MuCGEC} \textit{test}}                                             & \multicolumn{3}{c}{\textbf{FCGEC} \textit{test}}     \\
\textbf{Single System}                  & \textbf{Parameters}                                    & \textbf{P}     & \textbf{R}     & $\mathbf{F_{0.5}}$                                          & \textbf{P}     & \textbf{R}     & $\mathbf{F_{0.5}}$                                          & \textbf{P}     & \textbf{R}     & $\mathbf{F_{0.5}}$                                          & \textbf{P}     & \textbf{R}     & $\mathbf{F_{0.5}}$ \\ \hline
\multicolumn{14}{c}{\textbf{Primary Results}}                                                                                                                                                                                                                                                                                                                                                                                                           \\ \hline
\textbf{GECToR}                         & \multicolumn{1}{l|}{110M}                              & \textbf{77.5}  & 40.1           & \multicolumn{1}{c|}{65.3}                                   & \textbf{79.2}  & 53.9           & \multicolumn{1}{c|}{72.4}                                   & 46.72          & 27.14          & \multicolumn{1}{c|}{40.83}                                  & 46.11          & 34.35          & 43.16              \\
\textbf{BART}                           & \multicolumn{1}{l|}{400M}                              & 73.6           & 48.6           & \multicolumn{1}{c|}{66.7}                                   & 74.0           & {\ul 64.9}     & \multicolumn{1}{c|}{72.0}                                   & 41.90          & 29.48          & \multicolumn{1}{c|}{38.64}                                  & 38.38          & 37.62          & 38.23              \\
\textbf{T5}                             & \multicolumn{1}{l|}{770M}                              & -              & -              & \multicolumn{1}{c|}{66.1}                                   & -              & -              & \multicolumn{1}{c|}{72.1}                                   & -              & -              & \multicolumn{1}{c|}{-}                                      & -              & -              & -                  \\
\textbf{SynGEC}                         & \multicolumn{1}{l|}{110M+400M}                         & 74.7           & 49.0           & \multicolumn{1}{c|}{67.6}                                   & 75.1           & \textbf{65.5}  & \multicolumn{1}{c|}{72.9}                                   & \textbf{54.69} & 29.10          & \multicolumn{1}{c|}{\textbf{46.51}}                         & -              & -              & -                  \\ \hdashline
\textbf{SpanDC}                         & \multicolumn{1}{l|}{125M+209M}                         & 72.6           & 37.2           & \multicolumn{1}{c|}{61.0}                                   & 70.4           & 55.9           & \multicolumn{1}{c|}{66.9}                                   & -              & -              & \multicolumn{1}{c|}{-}                                      & -              & -              & -                  \\
\textbf{Multi-Encoder}                  & \multicolumn{1}{l|}{110M+107M}                         & 71.3           & 44.3           & \multicolumn{1}{c|}{63.5}                                   & 73.3           & 61.5           & \multicolumn{1}{c|}{70.6}                                   & -              & -              & \multicolumn{1}{c|}{-}                                      & -              & -              & -                  \\
\textbf{GEC-DePenD}                     & \multicolumn{1}{l|}{253M}                              & 73.2           & 37.8           & \multicolumn{1}{c|}{61.6}                                   & 72.9           & 53.2           & \multicolumn{1}{c|}{67.9}                                   & -              & -              & \multicolumn{1}{c|}{-}                                      & -              & -              & -                  \\
\textbf{TemplateGEC}                    & \multicolumn{1}{l|}{125M+770M}                         & 74.8           & \textbf{50.0}  & \multicolumn{1}{c|}{\textbf{68.1}}                          & 76.8           & 64.8           & \multicolumn{1}{c|}{{\ul 74.1}}                             & -              & -              & \multicolumn{1}{c|}{-}                                      & -              & -              & -                  \\
\rowcolor[HTML]{EFEFEF} 
\textbf{DeGLM-CoGLM}                    & \multicolumn{1}{l|}{\cellcolor[HTML]{EFEFEF}335M+335M} & 75.1           & 49.0           & \multicolumn{1}{c|}{\cellcolor[HTML]{EFEFEF}67.8}           & 76.4           & 63.4           & \multicolumn{1}{c|}{\cellcolor[HTML]{EFEFEF}73.4}           & {\ul 47.22}    & {\ul 30.08}    & \multicolumn{1}{c|}{\cellcolor[HTML]{EFEFEF}{\ul 42.39}}    & {\ul 52.95}    & \textbf{39.20} & {\ul 49.48}        \\
\rowcolor[HTML]{EFEFEF} 
\textbf{DeCoGLM}                        & \multicolumn{1}{l|}{\cellcolor[HTML]{EFEFEF}335M}      & {\ul 75.1}     & {\ul 49.4}     & \multicolumn{1}{c|}{\cellcolor[HTML]{EFEFEF}{\ul 68.0}}     & {\ul 77.4}     & 64.6           & \multicolumn{1}{c|}{\cellcolor[HTML]{EFEFEF}\textbf{74.4}}  & 45.01          & \textbf{31.77} & \multicolumn{1}{c|}{\cellcolor[HTML]{EFEFEF}41.55}          & \textbf{55.75} & {\ul 37.91}    & \textbf{50.96}     \\ \hline
\multicolumn{14}{c}{\textbf{Resource-restricted LLMs}}                                                                                                                                                                                                                                                                                                                                                                                                  \\ \hline
\textbf{ChatGLM2}                       & \multicolumn{1}{l|}{6B}                                & 61.72          & 45.58          & \multicolumn{1}{c|}{57.64}                                  & 56.89          & 58.73          & \multicolumn{1}{c|}{57.25}                                  & 31.35          & 21.39          & \multicolumn{1}{c|}{28.68}                                  & 44.30          & 17.08          & 33.59              \\
\textbf{ChatGLM3}                       & \multicolumn{1}{l|}{6B}                                & 60.63          & 47.50          & \multicolumn{1}{c|}{57.46}                                  & 59.48          & 60.37          & \multicolumn{1}{c|}{59.65}                                  & 30.62          & 21.60          & \multicolumn{1}{c|}{28.26}                                  & 41.06          & 19.93          & 33.88              \\
\textbf{LLaMA2/Baichuan}                & \multicolumn{1}{l|}{7B}                                & 67.24          & 51.84          & \multicolumn{1}{c|}{63.47}                                  & 66.16          & 66.12          & \multicolumn{1}{c|}{66.15}                                  & 36.47          & 25.18          & \multicolumn{1}{c|}{33.47}                                  & 51.83          & 24.08          & 42.12              \\
\textbf{LLaMA2/Baichuan}                & \multicolumn{1}{l|}{13B}                               & {\ul 68.43}    & \textbf{55.30} & \multicolumn{1}{c|}{{\ul 65.33}}                            & {\ul 69.46}    & \textbf{69.28} & \multicolumn{1}{c|}{{\ul 69.42}}                            & {\ul 37.91}    & {\ul 26.90}    & \multicolumn{1}{c|}{{\ul 35.04}}                            & \textbf{56.65} & {\ul 27.11}    & {\ul 46.52}        \\
\rowcolor[HTML]{EFEFEF} 
\textbf{DeGLM-CoGLM}                    & \multicolumn{1}{l|}{\cellcolor[HTML]{EFEFEF}335M+10B}  & \textbf{70.58} & {\ul 52.65}    & \multicolumn{1}{c|}{\cellcolor[HTML]{EFEFEF}\textbf{66.08}} & \textbf{72.80} & {\ul 67.57}    & \multicolumn{1}{c|}{\cellcolor[HTML]{EFEFEF}\textbf{71.69}} & \textbf{47.48} & \textbf{29.92} & \multicolumn{1}{c|}{\cellcolor[HTML]{EFEFEF}\textbf{42.49}} & {\ul 56.09}    & \textbf{38.02} & \textbf{51.22}     \\ \hline
\multicolumn{14}{c}{\textbf{GPT-4  Zeroshot}}                                                                                                                                                                                                                                                                                                                                                                                                           \\ \hline
\textbf{ZeroShot}                       & \multicolumn{1}{l|}{-}                                 & 59.64          & \textbf{58.32} & \multicolumn{1}{c|}{59.37}                                  & 55.69          & \textbf{70.44} & \multicolumn{1}{c|}{58.13}                                  & \textbf{36.36} & 27.71          & \multicolumn{1}{c|}{\textbf{34.22}}                         & 18.83          & 4.08           & 10.93              \\
\rowcolor[HTML]{EFEFEF} 
\cellcolor[HTML]{EFEFEF}\textbf{+DeGLM} & \multicolumn{1}{l|}{\cellcolor[HTML]{EFEFEF}-}         & \textbf{66.40} & 54.81          & \multicolumn{1}{c|}{\cellcolor[HTML]{EFEFEF}\textbf{63.70}} & \textbf{64.92} & 69.42          & \multicolumn{1}{c|}{\cellcolor[HTML]{EFEFEF}\textbf{65.78}} & 32.68          & \textbf{30.90} & \multicolumn{1}{c|}{\cellcolor[HTML]{EFEFEF}32.31}          & \textbf{25.60} & \textbf{16.98} & \textbf{23.24}     \\ \hline
\end{tabular}
}
\caption{
Results on English and Chinese GEC benchmarks. The parameter counts of the backbones of each system are shown in the second column. Under restricted resource, LLMs are fine-tuned using smaller datasets by LoRA. The highest metric is indicated in bold, while the second highest metric value is underlined.
}
\label{main-results}
\end{table*}

\subsection{Datasets and Evaluation}
\label{sec:data-and-evaluation}
For the English GEC task, we evaluate the performance on the CoNLL-14 test set \citep{ng-etal-2014-conll14} using the $M^2$ Scorer \cite{dahlmeier-ng-2012-m2scorer}, and on the BEA-19 test set \citep{bryant-etal-2019-bea-19} using the ERRANT scorer \citep{bryant-etal-2017-errant}. The model is pretrained on synthetic dataset C4-200M \citep{stahlberg-kumar-2021-synthetic-tag-corrupted} and fine-tuned on the cleaned Lang8 dataset (CLang8) \citep{rothe2021clang8_gec}. For the large version of CoGLM model, we utilize smaller datasets including FCE \citep{yannakoudakis-etal-2011-fce}, NUCLE \citep{dahlmeier-etal-2013-nucle}, and W\&I+LOCNESS \citep{bryant-etal-2019-bea-19} for fine-tuning, following \citet{zhou-etal-2023-improving-seq2seq-decoding-interventions}. The BEA-19 dev set is used for model selection.

For the Chinese GEC task, we synthesize pretraining data from the People's Daily corpus\footnote{\href{https://github.com/shibing624/pycorrector}{https://github.com/shibing624/pycorrector}} using rule-based insertion, replacement, and deletion. The models are fine-tuned on the Chinese Lang8 dataset \citep{zhao2018nlpcc-gec} and the HSK dataset, following \citet{zhang-etal-2022-mucgec}, and on the FCGEC training set, respectively. The models are evaluated on MuCGEC and FCGEC test sets using ChERRANT \citep{zhang-etal-2022-mucgec, xu-etal-2022-fcgec}. Further details are provided in Appendix \ref{sec:appendix-dataset}.

\subsection{Model Settings}
\noindent \textbf{Proposed Models} \quad
The open-source GLMs are utilized as the backbones for both DeCoGLM and separate models. The detection head comprises a feed-forward network with a single hidden layer, the dimension of which matches that of the GLM hidden state. The English base model employs \href{https://huggingface.co/THUDM/glm-roberta-large}{\texttt{glm-roberta-large}}, while \href{https://huggingface.co/THUDM/glm-large-chinese}{\texttt{glm-large-chinese}} is used as the Chinese base model. The large CoGLM models for error correction, denoted as CoGLM (10B), uses \href{https://huggingface.co/THUDM/glm-10b}{\texttt{glm-10b}} and \href{https://huggingface.co/THUDM/glm-10b-chinese}{\texttt{glm-10b-chinese}} as backbones. Due to the restriction of computational resources, large models are fine-tuned on the relatively small fine-tuning dataset mentioned in Section \ref{sec:data-and-evaluation} by LoRA \citep{hu2021lora}, without datasets for pretraining. Refer to Appendix \ref{sec:appendix-train-settings} for detailed configurations.

\noindent \textbf{Comparison with Previous Works} \quad 
In the main experiment, we present the results of single systems trained on parallel data without reranker. GECToR \citep{omelianchuk2020gector_gec} represents the Seq2Edit models, while BART and T5 \citep{2019bart, 2020t5} are SOTA backbones of Seq2Seq GEC methods. SynGEC \citep{zhang-etal-2022-syngec} incorporates syntactic information into the BART model. The performance of GECToR and BART model on the Chinese dataset is the reproduced result under our data configuration, and the results for BART on the English dataset are reported by \citet{zhang-etal-2022-syngec}. We also present the results of four models involving the detection-correction process. SpanDC \citep{chen-etal-2020-seq2span_gec} comprises a span detector and a generator. Multi-Encoder \citep{yuan-etal-2021-two_systems_detection_correction_gec} encodes error categories as auxiliary information. GEC-DePend \citep{yakovlev-etal-2023-gec-depend} integrates error detection with correction by the MLM. TemplateGEC \cite{li-etal-2023-templategec} uses the output of the GECToR model as supplementary information for Seq2Seq models. 

\noindent \textbf{Comparison with LLMs} \quad For the LLMs treating GEC as a Seq2Seq task, we fine-tune ChatGLM2, ChatGLM3 \citep{du2021glm}, and Llama2 \citep{touvron2023llama2} with LoRA. As Llama2 is not optimized for Chinese, the results on the Chinese dataset are obtained using the Baichuan \citep{yang2023baichuan2} models.

\noindent \textbf{GPT-4} \quad We report the zero-shot performance of GPT-4 on four datasets with prompting. We attempt to incorporate detection results in the form of masked text into the prompt of GPT-4, aiming to enhance the performance on GEC tasks.

\subsection{Main Results}
\label{sec:main-results}

Table \ref{main-results} presents the main results. According to the last two rows of primary results, the integrated detection-correction model outperforms the separate models in most cases in terms of the $F_{0.5}$ metric, despite having only half the parameter count. This suggests that the designed multi-task learning mutually reinforces detection and correction, which will be further discussed in Section \ref{sec:mutual-benefits}. DeCoGLM achieves the highest or second-highest $F_{0.5}$ performance on three datasets, demonstrating comparable performance to SOTA GEC models. Considering the model parameter counts, our model outperforms all previous works with the detection-correction process, indicating that the well-designed detection-correction structure can achieve the SOTA level in GEC, a field typically dominated by Seq2Seq models. Furthermore, the inference speed of the localized error correction is significantly faster than the globalized error correction of the Seq2Seq method, with the details provided in Appendix \ref{sec:inference-speed}. These results also underscore the potential of GLM in the GEC field.

Despite limitations of data quantity and fine-tuning methods, fine-tuning LLMs with over 10B parameters yields results approaching SOTA level, suggesting that LLMs can reduce the need for extensive supervised data for fine-tuning. The strategy of small detection models assisting large models in localized correction yields improved performance across all datasets, primarily due to higher precision. This suggests that the model reduces over-correction at the expense of a certain level of recall. On the English dataset, GPT-4 exhibits a similar trend when incorporated with detection results, indicating that detection results can stably improve the GEC capability of LLMs, thus presenting a promising future direction for GEC.

\section{Analysis}

\subsection{Interaction of Detection and Correction}
\label{sec:mutual-benefits}
\begin{table*}[ht]
\centering
\resizebox{\textwidth}{!}{
\begin{tabular}{lc|cccc|cccc|cccc}
\hline
\textbf{}        &                     & \multicolumn{4}{c|}{\textbf{BEA-19} \textit{dev}}                         & \multicolumn{4}{c|}{\textbf{MuCGEC} \textit{dev}}                         & \multicolumn{4}{c}{\textbf{FCGEC} \textit{dev}}                           \\
\textbf{Model}   & \textbf{Parameters} & $\mathbf{Acc_D}$ & $\mathbf{Rec_E}$ & $\mathbf{Rec_I}$ & $\mathbf{Acc_C}$ & $\mathbf{Acc_D}$ & $\mathbf{Rec_E}$ & $\mathbf{Rec_I}$ & $\mathbf{Acc_C}$ & $\mathbf{Acc_D}$ & $\mathbf{Rec_E}$ & $\mathbf{Rec_I}$ & $\mathbf{Acc_C}$ \\ \hline
\textbf{DeCoGLM} & 335M                & \textbf{94.56}   & \textbf{65.60}   & \textbf{63.95}   & \textbf{90.54}   & \textbf{84.44}   & \textbf{52.72}   & \textbf{25.74}   & 74.24            & 96.74            & 54.57            & \textbf{51.13}   & \textbf{84.44}   \\
\textbf{DeGLM}   & 335M                & 94.49            & 64.88            & 62.82            & -                & 84.20            & 52.54            & 23.93            & -                & \textbf{96.96}   & \textbf{54.85}   & 47.87            & -                \\
\textbf{CoGLM}   & 335M                & -                & -                & -                & 90.27            & -                & -                & -                & \textbf{74.55}   & -                & -                & -                & 83.94            \\ \hline
\end{tabular}
}
\caption{
The metrics of detection and correction tasks on the development set. The results are presented using four metrics (3 detection metrics and 1 correction metric): overall accuracy in the detection phase ($\mathbf{Acc_D}$), recall for the detection label ERROR ($\mathbf{Rec_E}$), recall for the detection label INSERT ($\mathbf{Rec_I}$), and the accuracy of next token prediction during the localized error correction.
}
\label{tab:collaboration-study}
\end{table*}

In Section \ref{sec:main-results}, we mentioned that detection and correction tasks can mutually benefit each other. To further verify this, we conduct experiments using the integrated model (DeCoGLM, 335M) and two separate models for detection and correction (DeGLM, 335M; CoGLM, 335M), as shown in the primary results in Table \ref{main-results}. Without employing two-stage fine-tuning involving data enhancement, the models are trained on the same dataset, and their performance on the development set for detection and correction metrics is presented in Table \ref{tab:collaboration-study}. 
The integrated model exhibits superior detection and correction capabilities over separate models. A fairer comparison should involve two separate models with a parameter count of $335\mathrm{M}/2=167.5\mathrm{M}$, but currently, there is no GLM backbone of approximately this size. In this scenario with fewer parameters, the advantage of the integrated model is expected to be even greater.

\subsection{Weights of Multi-Task Training}
\label{sec:weight-study}
\begin{table}
\centering
\resizebox{0.8\columnwidth}{!}{
\begin{tabular}{c|c|ccc}
\hline
\textbf{}                    & \textbf{}     & \multicolumn{3}{c}{$\mathbf{F_{0.5}}$ \textbf{on dev set}} \\
$w_D$                        & $\alpha_{EI}$ & \textbf{BEA-19}    & \textbf{MuCGEC}    & \textbf{FCGEC}   \\ \hline
\textbf{20}                  & \textbf{2}    & 60.30              & 34.45              & 40.57            \\ \hline
\multirow{5}{*}{\textbf{10}} & -             & 60.09              & {\ul 35.17}        & 41.52            \\
                             & \textbf{1}    & 59.93              & 34.25              & \textbf{42.89}   \\
                             & \textbf{2}    & \textbf{60.81}     & 35.09              & {\ul 42.49}      \\
                             & \textbf{3}    & 60.29              & \textbf{35.82}     & 40.72            \\
                             & \textbf{4}    & 60.12              & 35.03              & 41.49            \\ \hline
\textbf{5}                   & \textbf{2}    & {\ul 60.60}        & 34.53              & 42.10            \\ \hline
\textbf{1}                   & \textbf{2}    & 59.64              & 33.23              & 36.72            \\ \hline
\end{tabular}
}
\caption{
The preliminary experimental results of different loss weights. $w_D$ and $\alpha_{EI}$ is defined in Section \ref{sec:multi-task-organization} and \ref{sec:error-detection}. The "-" value of $\alpha_{EI}$ represents the usage of cross-entropy other than Focal Loss.
}
\label{weight-study}
\end{table}


To establish two weights that significantly impact the training objective: the detection loss weight \(w_D\) in Equation \ref{eq:loss}, and the ERROR and INSERT loss weight \(\alpha_{EI}\) in Equation \ref{eq:detection-loss}, we conduct preliminary experiments, which include only the two stages of fine-tuning. The obtained results are presented in Table \ref{weight-study}. Based on a preliminary observation on the loss scale, we initially set \(w_D=10\) and explore experimental results under varying \(\alpha_{EI}\). The outcomes suggest that the Focal Loss along with moderately increasing \(\alpha_{EI}\) to achieve aggressive detection introduced in Section \ref{sec:multi-task-organization} is effective. After setting \(\alpha_{EI}=2\), we conducted additional experiments with different \(w_D\). The overall experimental results indicate that \(\alpha_{EI}=2\) and \(w_D=10\) constitute a suitable setup.

\begin{table}
\centering
\resizebox{\columnwidth}{!}{
\begin{tabular}{cccc|cccccc}
\hline
\textbf{}    &              &              &              & \multicolumn{3}{c}{\textbf{CoNLL-14} \textit{test}}  & \multicolumn{3}{c}{\textbf{BEA-19} \textit{test}}    \\
\textbf{K}   & \textbf{E}   & \textbf{I}   & \textbf{D}   & \textbf{P}     & \textbf{R}     & $\mathbf{F_{0.5}}$ & \textbf{P}     & \textbf{R}     & $\mathbf{F_{0.5}}$ \\ \hline
$\checkmark$ & $\checkmark$ &              &              & \textbf{69.67} & 50.91          & \textbf{64.89}     & 72.18          & 65.14          & 70.65              \\
$\checkmark$ & $\checkmark$ & $\checkmark$ &              & 69.25          & \textbf{51.26} & 64.71              & \textbf{72.33} & \textbf{65.46} & \textbf{70.85}     \\
$\checkmark$ & $\checkmark$ & $\checkmark$ & $\checkmark$ & 68.48          & 49.95          & 63.75              & 71.23          & 64.48          & 69.77              \\ \hline
\end{tabular}
}
\caption{\label{template-study}
Results under different detection label sets. \textbf{K}=KEEP, \textbf{E}=ERROR, \textbf{I}=INSERT and \textbf{D}=DELETE.
}
\end{table}
\subsection{Detection Label Set}
In the design outlined in Section \ref{sec:error-detection}, ERROR includes both replacement and deletion, as the deletion can be considered as replacing with zero-length text. The results for this design are shown in the second row of Table \ref{template-study}. INSERT can also be further merged into the ERROR label. This can be achieved by considering the INSERT operation as replacing the token \(x_i\) at the insertion position with \(x_i \boldsymbol{c_j}\), where \(\boldsymbol{c_j}\) represents tokens to be inserted. The results corresponding to this approach are shown in the first row of Table \ref{template-study}. Additionally, we demonstrate the results of applying four detection labels (KEEP, ERROR, INSERT, DELETE) in the last row. Overall, our designed three-label scheme performs relatively better, as the insertion operation in the two-label mode requires disrupting the correct part of the source text, and encountering DELETE in the four-label mode will lead to direct deletion, which makes the model unable to recover from faults in the error correction phase.

\subsection{Ablation Study}
\label{sec:ablation-study}
\begin{table*}
\centering
\small
\begin{tabular}{cc|ccc|cccccc}
\hline
                  &                     & \textbf{}     &               &               & \multicolumn{3}{c}{\textbf{CoNLL-14} \textit{test}}  & \multicolumn{3}{c}{\textbf{BEA-19} \textit{test}}    \\
\textbf{BackBone} & \textbf{Pretrained} & \textbf{SFT1} & \textbf{SFT2} & \textbf{Ctrl} & \textbf{P}     & \textbf{R}     & $\mathbf{F_{0.5}}$ & \textbf{P}     & \textbf{R}     & $\mathbf{F_{0.5}}$ \\ \hline
GLM-Roberta       & Yes                 & $\checkmark$  & $\checkmark$  & $\checkmark$  & 75.07          & 49.40          & \textbf{68.00}     & \textbf{77.36} & 64.63          & \textbf{74.43}     \\
GLM-Roberta       & Yes                 & $\checkmark$  & $\checkmark$  & $\times$      & 70.47          & 54.96          & 66.70              & 72.75          & 69.28          & 72.03              \\
GLM-Roberta       & Yes                 & $\checkmark$  & $\times$      & $\checkmark$  & \textbf{75.27} & 48.24          & 67.69              & 76.55          & 62.34          & 73.21              \\
GLM-Roberta       & Yes                 & $\checkmark$  & $\times$      & $\times$      & 68.38          & \textbf{57.35} & 65.84              & 69.00          & \textbf{71.02} & 69.39              \\
GLM-Roberta       & Yes                 & $\times$      & $\times$      & $\times$      & 54.04          & 45.99          & 52.21              & 45.12          & 58.60          & 47.30              \\ \hline
GLM-Roberta       & No                  & $\checkmark$  & $\checkmark$  & $\checkmark$  & \textbf{72.78} & 46.42          & \textbf{65.36}     & \textbf{75.54} & 59.87          & \textbf{71.78}     \\
GLM-Roberta       & No                  & $\checkmark$  & $\checkmark$  & $\times$      & 69.25          & 51.26          & 64.71              & 72.33          & 65.46          & 70.85              \\
GLM-Roberta       & No                  & $\checkmark$  & $\times$      & $\checkmark$  & 68.25          & 49.33          & 63.39              & 69.66          & 61.94          & 67.97              \\
GLM-Roberta       & No                  & $\checkmark$  & $\times$      & $\times$      & 63.92          & \textbf{52.46} & 61.25              & 66.27          & \textbf{66.01} & 66.21              \\ \hline
BART-large        & No                  & $\checkmark$  & $\checkmark$  & $\checkmark$  & \textbf{69.53} & 45.62          & \textbf{62.93}     & \textbf{72.01} & 57.84          & \textbf{68.64}     \\
BART-large        & No                  & $\checkmark$  & $\checkmark$  & $\times$      & 66.39          & 49.80          & 62.24              & 69.25          & 63.28          & 67.97              \\
BART-large        & No                  & $\checkmark$  & $\times$      & $\checkmark$  & 67.54          & 43.66          & 60.88              & 68.08          & 55.14          & 65.03              \\
BART-large        & No                  & $\checkmark$  & $\times$      & $\times$      & 62.75          & \textbf{50.40} & 59.81              & 64.67          & \textbf{63.62} & 64.46              \\ \hline
\end{tabular}
\caption{\label{ablation-study}
Ablation study results. The "Ctrl" denotes the proposed detection control. 
}
\end{table*}

To explore the effectiveness of various components in the designed detection-correction model, we conduct an ablation study focusing on synthetic data, backbone, two-stage fine-tuning, and detection control. The results are shown in Table \ref{ablation-study}. 

\noindent \textbf{Effectiveness of synthetic data} \quad In the proposed model, both the English and Chinese models undergo pretraining with a large-scale synthetic dataset of GEC. A comparison between the top and middle rows of Table \ref{ablation-study} reveals that pretraining indeed provides a stable improvement in model performance, although the data used is not from real scenarios.
 
\noindent \textbf{Effectiveness of GLM backbone} \quad The detection-correction structure can also be implemented in Seq2Seq models. We applied the proposed method to the BART model and conducted experiments. An additional detection head is integrated into the BART encoder, while the decoder generates text pieces for localized error correction. The experimental results, depicted in the bottom rows of Table \ref{ablation-study}, consistently demonstrate superior performance when employing GLM as the backbone compared to using BART. This can be attributed, in part, to the consistency between the original pretraining task of GLM and the training objective of the correction task, as defined in Equation \ref{eq:glm-loss}. However, the pretraining pattern of BART differs. Additionally, the separation of BART's encoder and decoder into two distinct modules may not effectively foster the mutual enhancement of detection and correction abilities in multi-task learning.

\noindent \textbf{Effectiveness of Two Stage Fine-tuning} \quad As described in Section \ref{sec:multi-task-organization}, two fine-tuning stages differ in the training data: SFT1 constructs training samples using only ground-truth detection labels, while SFT2 utilizes both ground-truth detection labels and the detection results from the model trained in the first stage. As evident from the comparison in Table \ref{ablation-study}, SFT1 significantly improves the model's performance than the model pretrained on the synthetic dataset. Comparing the results exclusively differing in SFT2 in Table \ref{ablation-study}, it is observed that SFT2 consistently enhances $F_{0.5}$, primarily attributed to the improvement in precision while maintaining recall relatively constant. This validates the effectiveness of the two-stage supervised fine-tuning design.

\noindent \textbf{Detection Control} \quad From Table \ref{ablation-study}, it is evident that, under the scenario of employing the same trained model, setting three hyper-parameters for the detection phase also enhances the $F_{0.5}$ performance. This approach primarily aims at improving precision. However, upon closer inspection, it is noticeable that this technique results in a more substantial reduction in recall compared to the second-stage fine-tuning. For all GLM models incorporating detection control, the recall on the CoNLL-14 test set is consistently below 50\%, and the recall on the BEA-19 test set is consistently below 65\%. Thus, the effectiveness of detection control stems more from the trade-off between precision and recall, as discussed in the next section.

\subsection{Precision-Recall Trade off}
\label{sec:pre-rec-trade-off}

\begin{figure}[ht]
    \centering
    \includegraphics[width=1\columnwidth]{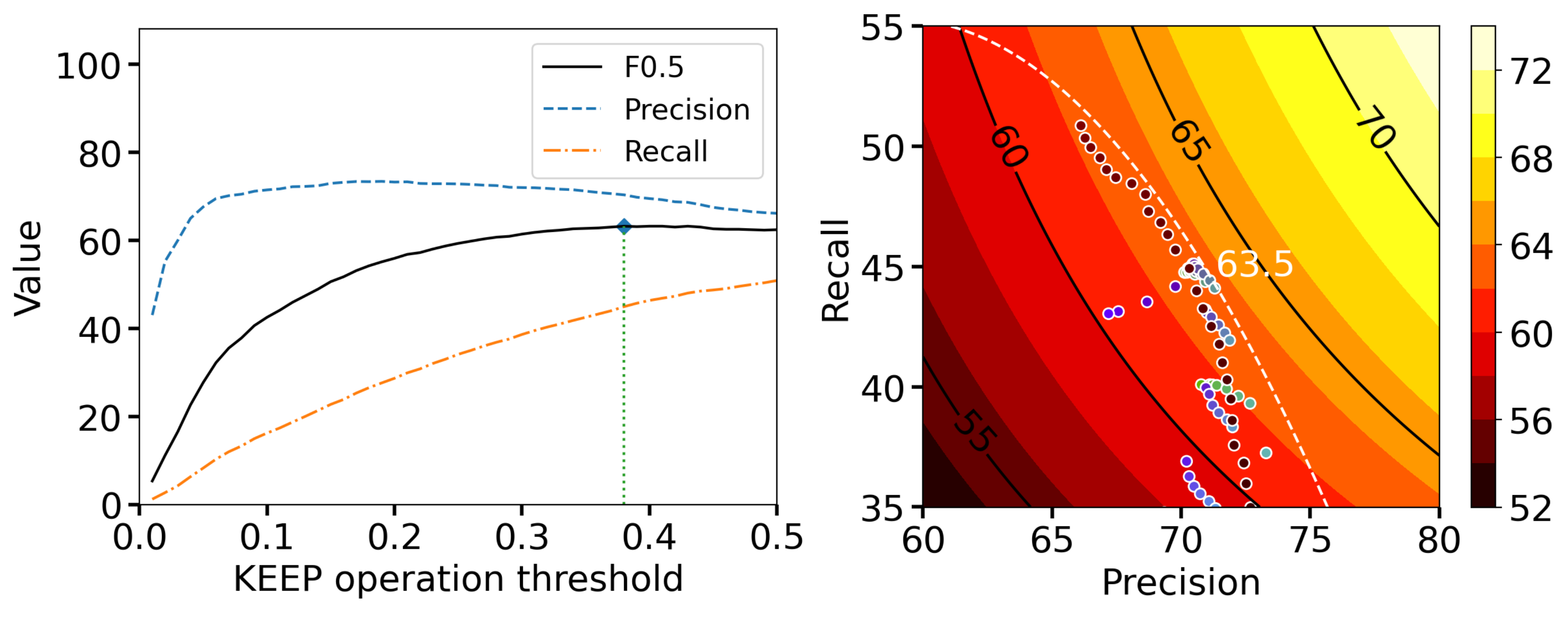}
    \caption{Results of detection control on BEA-19 dev set. The heat value represents the value of $F_{0.5}$.}
    \label{fig:ctrl}
\end{figure}

Adjusting the threshold for KEEP prediction probability (\(\delta\)) and the probability lower bounds for ERROR and INSERT predictions (\(\phi_e, \phi_i\)) defined in Section \ref{sec:detection-control} allows for further adjustment of precision and recall, resulting in improved \(F_{0.5}\) scores. We performed a parameter search on the validation set to identify configurations maximizing \(F_{0.5}\), and the results are depicted in Figure \ref{fig:ctrl}.

Without setting \(\phi_e\) and \(\phi_i\), \(\delta=0.38\) achieved the highest \(F_{0.5}\) of 63.2 on BEA-19 dev set. Then, we fix \(\delta=0.38\) and perform a grid search for \(\phi_e, \phi_i\). All results are presented as points in the right plot of Figure \ref{fig:ctrl}, and nearly all points are located within the region enclosed by the dashed line in the bottom-left. The dashed line represents the boundary of the model's capability, and the intersection point with the $F_{0.5}$ contour line represents the optimal performance attainable by the model. The point with the highest \(F_{0.5}=63.5\) is the one closest to the intersection point, with \(\phi_e=0.5\) and \(\phi_i=0.6\). Under this parameter configuration, the model achieved an \(F_{0.5}\) value of 74.43 on the BEA-19 test set, as shown in Table \ref{main-results}. The detection control offers such a straightforward implementation of the precision-recall trade-off.

\section{Conclusion}

We introduce a novel language-agnostic detection-correction structure via GLM for the GEC task. The structure employs a three-label error detection pattern and uses Focal Loss for aggressive detection. The correction phase leverages the mask-infilling capability of GLM to generate correct text pieces. A multi-task learning approach is designed to integrate both functionalities within the same model, optimized using a weighted loss function. Experimental results show proposed model DeCoGLM outperforms previous detection-correction structures and achieves $F_{0.5}$ scores comparable to SOTA on English and Chinese GEC benchmarks. The effectiveness of the detection-correction structure is further validated by applying it to open-source LLMs and GPT-4, indicating that incorporating error detection information improves the performance of LLMs on GEC datasets by reducing over-correction. Ablation studies confirm the efficacy of our model design and the ability to trade off precision and recall can be realized by detection control. We aim for this work to further guide GEC research within the detection-correction paradigm. The code and related models are available at \href{https://github.com/GMago-LeWay/GECFramework}{https://github.com/GMago-LeWay/GECFramework}.

\section*{Limitations}
Incremental methods proven effective on Seq2Seq models, such as incorporating syntactic information \citep{zhang-etal-2022-syngec}, refining training data \citep{mita-etal-2020-noise-refinement}, and employing additional models for reranking during the generation phase \citep{zhang-etal-2023-transformer-reranker, zhou-etal-2023-improving-seq2seq-decoding-interventions}, are not implemented in this work. The main objective of this paper is to propose a novel GEC architecture, with these additional tricks serving as potential avenues for future extensions. Furthermore, due to resource restrictions, we are unable to apply our integrated detection-correction structure to LLMs. This is because the sequence labeling task differs from the generative tasks that LLMs are designed to perform, necessitating full-parameter fine-tuning to integrate the two tasks. Additionally, in our investigation of LLMs as correction models, models with parameters exceeding 13B are not utilized. The absence of full-parameter fine-tuning on LLMs and experiments with larger models due to resource constraints leaves room for further exploration of the application of the detection-correction paradigm on LLMs.

\section*{Ethics Statement}
The datasets and models we used are publicly available and utilized only for research purposes. The datasets do not contain any information that names or uniquely identifies individual people or offensive content. LLMs are utilized in our experiments, consistent with their intended use in natural language processing tasks. The models we designed will be published and intended for academic research in the field of grammatical error correction, in accordance with the original access conditions of the models used.

The detection-correction structure we designed limits the model to making only localized modifications to the text, preventing it from generating text without constraints, thereby significantly reducing the potential risks associated with the model. However, It is worth noting that the modifications made by the designed model may alter certain facts in the text, leading to hallucination, especially when modifications occur in named entities.

ChatGPT is utilized as the AI Assistant to polish the paper writing.

\section*{Acknowledgements}
We thank all the reviewers for their valuable comments to improve our paper. 
This work was supported by National Natural Science Foundation of China (62036001). The corresponding author is Houfeng Wang.


\bibliography{custom}

\begin{thebibliography}{65}
\expandafter\ifx\csname natexlab\endcsname\relax\def\natexlab#1{#1}\fi

\bibitem[{Awasthi et~al.(2019)Awasthi, Sarawagi, Goyal, Ghosh, and Piratla}]{awasthi2019parallel_edit_seq2edit_gec}
Abhijeet Awasthi, Sunita Sarawagi, Rasna Goyal, Sabyasachi Ghosh, and Vihari Piratla. 2019.
\newblock \href {https://doi.org/10.18653/v1/D19-1435} {Parallel iterative edit models for local sequence transduction}.
\newblock In \emph{Proceedings of the 2019 Conference on Empirical Methods in Natural Language Processing and the 9th International Joint Conference on Natural Language Processing (EMNLP-IJCNLP)}, pages 4260--4270, Hong Kong, China. Association for Computational Linguistics.

\bibitem[{Bell et~al.(2019)Bell, Yannakoudakis, and Rei}]{bell2019context_detection_gec}
Samuel Bell, Helen Yannakoudakis, and Marek Rei. 2019.
\newblock \href {https://doi.org/10.18653/v1/W19-4410} {Context is key: Grammatical error detection with contextual word representations}.
\newblock In \emph{Proceedings of the Fourteenth Workshop on Innovative Use of NLP for Building Educational Applications}, pages 103--115, Florence, Italy. Association for Computational Linguistics.

\bibitem[{Bout et~al.(2023)Bout, Podolskiy, Nikolenko, and Piontkovskaya}]{bout-etal-2023-training-schedule}
Andrey Bout, Alexander Podolskiy, Sergey Nikolenko, and Irina Piontkovskaya. 2023.
\newblock \href {https://doi.org/10.18653/v1/2023.emnlp-main.355} {Efficient grammatical error correction via multi-task training and optimized training schedule}.
\newblock In \emph{Proceedings of the 2023 Conference on Empirical Methods in Natural Language Processing}, pages 5800--5816, Singapore. Association for Computational Linguistics.

\bibitem[{Brown et~al.(2020)Brown, Mann, Ryder, Subbiah, Kaplan, Dhariwal, Neelakantan, Shyam, Sastry, Askell et~al.}]{brown2020gpt3}
Tom Brown, Benjamin Mann, Nick Ryder, Melanie Subbiah, Jared~D Kaplan, Prafulla Dhariwal, Arvind Neelakantan, Pranav Shyam, Girish Sastry, Amanda Askell, et~al. 2020.
\newblock Language models are few-shot learners.
\newblock \emph{Advances in neural information processing systems}, 33:1877--1901.

\bibitem[{Bryant et~al.(2019)Bryant, Felice, Andersen, and Briscoe}]{bryant-etal-2019-bea-19}
Christopher Bryant, Mariano Felice, {\O}istein~E. Andersen, and Ted Briscoe. 2019.
\newblock \href {https://doi.org/10.18653/v1/W19-4406} {The {BEA}-2019 shared task on grammatical error correction}.
\newblock In \emph{Proceedings of the Fourteenth Workshop on Innovative Use of NLP for Building Educational Applications}, pages 52--75, Florence, Italy. Association for Computational Linguistics.

\bibitem[{Bryant et~al.(2017)Bryant, Felice, and Briscoe}]{bryant-etal-2017-errant}
Christopher Bryant, Mariano Felice, and Ted Briscoe. 2017.
\newblock \href {https://doi.org/10.18653/v1/P17-1074} {Automatic annotation and evaluation of error types for grammatical error correction}.
\newblock In \emph{Proceedings of the 55th Annual Meeting of the Association for Computational Linguistics (Volume 1: Long Papers)}, pages 793--805, Vancouver, Canada. Association for Computational Linguistics.

\bibitem[{Caines et~al.(2023)Caines, Benedetto, Taslimipoor, Davis, Gao, Andersen, Yuan, Elliott, Moore, Bryant et~al.}]{caines2023teaching_application_gec}
Andrew Caines, Luca Benedetto, Shiva Taslimipoor, Christopher Davis, Yuan Gao, Oeistein Andersen, Zheng Yuan, Mark Elliott, Russell Moore, Christopher Bryant, et~al. 2023.
\newblock On the application of large language models for language teaching and assessment technology.
\newblock \emph{arXiv preprint arXiv:2307.08393}.

\bibitem[{Chang et~al.(2024)Chang, Wang, Wang, Wu, Yang, Zhu, Chen, Yi, Wang, Wang, Ye, Zhang, Chang, Yu, Yang, and Xie}]{chang2023survey_eval_llm}
Yupeng Chang, Xu~Wang, Jindong Wang, Yuan Wu, Linyi Yang, Kaijie Zhu, Hao Chen, Xiaoyuan Yi, Cunxiang Wang, Yidong Wang, Wei Ye, Yue Zhang, Yi~Chang, Philip~S. Yu, Qiang Yang, and Xing Xie. 2024.
\newblock \href {https://doi.org/10.1145/3641289} {A survey on evaluation of large language models}.
\newblock \emph{ACM Trans. Intell. Syst. Technol.}
\newblock Just Accepted.

\bibitem[{Chen et~al.(2020)Chen, Ge, Zhang, Wei, and Zhou}]{chen-etal-2020-seq2span_gec}
Mengyun Chen, Tao Ge, Xingxing Zhang, Furu Wei, and Ming Zhou. 2020.
\newblock \href {https://doi.org/10.18653/v1/2020.emnlp-main.581} {Improving the efficiency of grammatical error correction with erroneous span detection and correction}.
\newblock In \emph{Proceedings of the 2020 Conference on Empirical Methods in Natural Language Processing (EMNLP)}, pages 7162--7169, Online. Association for Computational Linguistics.

\bibitem[{Choe et~al.(2019)Choe, Ham, Park, and Yoon}]{choe-etal-2019-transferlearning_gec}
Yo~Joong Choe, Jiyeon Ham, Kyubyong Park, and Yeoil Yoon. 2019.
\newblock \href {https://doi.org/10.18653/v1/W19-4423} {A neural grammatical error correction system built on better pre-training and sequential transfer learning}.
\newblock In \emph{Proceedings of the Fourteenth Workshop on Innovative Use of NLP for Building Educational Applications}, pages 213--227, Florence, Italy. Association for Computational Linguistics.

\bibitem[{Coyne et~al.(2023)Coyne, Sakaguchi, Galvan-Sosa, Zock, and Inui}]{coyne2023gpt-gec-en}
Steven Coyne, Keisuke Sakaguchi, Diana Galvan-Sosa, Michael Zock, and Kentaro Inui. 2023.
\newblock Analyzing the performance of gpt-3.5 and gpt-4 in grammatical error correction.
\newblock \emph{arXiv preprint arXiv:2303.14342}.

\bibitem[{Dahlmeier and Ng(2012)}]{dahlmeier-ng-2012-m2scorer}
Daniel Dahlmeier and Hwee~Tou Ng. 2012.
\newblock \href {https://aclanthology.org/N12-1067} {Better evaluation for grammatical error correction}.
\newblock In \emph{Proceedings of the 2012 Conference of the North {A}merican Chapter of the Association for Computational Linguistics: Human Language Technologies}, pages 568--572, Montr{\'e}al, Canada. Association for Computational Linguistics.

\bibitem[{Dahlmeier et~al.(2013)Dahlmeier, Ng, and Wu}]{dahlmeier-etal-2013-nucle}
Daniel Dahlmeier, Hwee~Tou Ng, and Siew~Mei Wu. 2013.
\newblock \href {https://aclanthology.org/W13-1703} {Building a large annotated corpus of learner {E}nglish: The {NUS} corpus of learner {E}nglish}.
\newblock In \emph{Proceedings of the Eighth Workshop on Innovative Use of {NLP} for Building Educational Applications}, pages 22--31, Atlanta, Georgia. Association for Computational Linguistics.

\bibitem[{Devlin et~al.(2018)Devlin, Chang, Lee, and Toutanova}]{2018bert}
Jacob Devlin, Ming{-}Wei Chang, Kenton Lee, and Kristina Toutanova. 2018.
\newblock \href {http://arxiv.org/abs/1810.04805} {{BERT:} pre-training of deep bidirectional transformers for language understanding}.
\newblock \emph{CoRR}, abs/1810.04805.

\bibitem[{Du et~al.(2022)Du, Qian, Liu, Ding, Qiu, Yang, and Tang}]{du2021glm}
Zhengxiao Du, Yujie Qian, Xiao Liu, Ming Ding, Jiezhong Qiu, Zhilin Yang, and Jie Tang. 2022.
\newblock \href {https://doi.org/10.18653/v1/2022.acl-long.26} {{GLM}: General language model pretraining with autoregressive blank infilling}.
\newblock In \emph{Proceedings of the 60th Annual Meeting of the Association for Computational Linguistics (Volume 1: Long Papers)}, pages 320--335, Dublin, Ireland. Association for Computational Linguistics.

\bibitem[{Fang et~al.(2023{\natexlab{a}})Fang, Hu, Wong, Wan, Chao, and Chang}]{fang-etal-2023-multimodal-gec}
Tao Fang, Jinpeng Hu, Derek~F. Wong, Xiang Wan, Lidia~S. Chao, and Tsung-Hui Chang. 2023{\natexlab{a}}.
\newblock \href {https://doi.org/10.18653/v1/2023.findings-acl.594} {Improving grammatical error correction with multimodal feature integration}.
\newblock In \emph{Findings of the Association for Computational Linguistics: ACL 2023}, pages 9328--9344, Toronto, Canada. Association for Computational Linguistics.

\bibitem[{Fang et~al.(2023{\natexlab{b}})Fang, Yang, Lan, Wong, Hu, Chao, and Zhang}]{fang2023chatgpt-gec-en}
Tao Fang, Shu Yang, Kaixin Lan, Derek~F Wong, Jinpeng Hu, Lidia~S Chao, and Yue Zhang. 2023{\natexlab{b}}.
\newblock Is chatgpt a highly fluent grammatical error correction system? a comprehensive evaluation.
\newblock \emph{arXiv preprint arXiv:2304.01746}.

\bibitem[{Fu et~al.(2023)Fu, Yang, So, Lam, Bing, and Collier}]{fu2023peft}
Zihao Fu, Haoran Yang, Anthony Man-Cho So, Wai Lam, Lidong Bing, and Nigel Collier. 2023.
\newblock \href {https://doi.org/10.1609/aaai.v37i11.26505} {On the effectiveness of parameter-efficient fine-tuning}.
\newblock \emph{Proceedings of the AAAI Conference on Artificial Intelligence}, 37(11):12799--12807.

\bibitem[{Ge et~al.(2018)Ge, Wei, and Zhou}]{ge-etal-2018-fluency_gec}
Tao Ge, Furu Wei, and Ming Zhou. 2018.
\newblock \href {https://doi.org/10.18653/v1/P18-1097} {Fluency boost learning and inference for neural grammatical error correction}.
\newblock In \emph{Proceedings of the 56th Annual Meeting of the Association for Computational Linguistics (Volume 1: Long Papers)}, pages 1055--1065, Melbourne, Australia. Association for Computational Linguistics.

\bibitem[{Grundkiewicz et~al.(2019)Grundkiewicz, Junczys-Dowmunt, and Heafield}]{grundkiewicz2019synthetic_data_gec}
Roman Grundkiewicz, Marcin Junczys-Dowmunt, and Kenneth Heafield. 2019.
\newblock \href {https://doi.org/10.18653/v1/W19-4427} {Neural grammatical error correction systems with unsupervised pre-training on synthetic data}.
\newblock In \emph{Proceedings of the Fourteenth Workshop on Innovative Use of NLP for Building Educational Applications}, pages 252--263, Florence, Italy. Association for Computational Linguistics.

\bibitem[{Hu et~al.(2021)Hu, Shen, Wallis, Allen-Zhu, Li, Wang, Wang, and Chen}]{hu2021lora}
Edward~J Hu, Yelong Shen, Phillip Wallis, Zeyuan Allen-Zhu, Yuanzhi Li, Shean Wang, Lu~Wang, and Weizhu Chen. 2021.
\newblock Lora: Low-rank adaptation of large language models.
\newblock \emph{arXiv preprint arXiv:2106.09685}.

\bibitem[{Junczys-Dowmunt et~al.(2018)Junczys-Dowmunt, Grundkiewicz, Guha, and Heafield}]{junczys2018first_transformer_for_gec}
Marcin Junczys-Dowmunt, Roman Grundkiewicz, Shubha Guha, and Kenneth Heafield. 2018.
\newblock \href {https://doi.org/10.18653/v1/N18-1055} {Approaching neural grammatical error correction as a low-resource machine translation task}.
\newblock In \emph{Proceedings of the 2018 Conference of the North {A}merican Chapter of the Association for Computational Linguistics: Human Language Technologies, Volume 1 (Long Papers)}, pages 595--606, New Orleans, Louisiana. Association for Computational Linguistics.

\bibitem[{Kaneko et~al.(2019)Kaneko, Hotate, Katsumata, and Komachi}]{kaneko2019transformer_reranking_for_gec}
Masahiro Kaneko, Kengo Hotate, Satoru Katsumata, and Mamoru Komachi. 2019.
\newblock \href {https://doi.org/10.18653/v1/W19-4422} {{TMU} transformer system using {BERT} for re-ranking at {BEA} 2019 grammatical error correction on restricted track}.
\newblock In \emph{Proceedings of the Fourteenth Workshop on Innovative Use of NLP for Building Educational Applications}, pages 207--212, Florence, Italy. Association for Computational Linguistics.

\bibitem[{Kaneko et~al.(2020)Kaneko, Mita, Kiyono, Suzuki, and Inui}]{kaneko-etal-2020-bert-fused-ged}
Masahiro Kaneko, Masato Mita, Shun Kiyono, Jun Suzuki, and Kentaro Inui. 2020.
\newblock \href {https://doi.org/10.18653/v1/2020.acl-main.391} {Encoder-decoder models can benefit from pre-trained masked language models in grammatical error correction}.
\newblock In \emph{Proceedings of the 58th Annual Meeting of the Association for Computational Linguistics}, pages 4248--4254, Online. Association for Computational Linguistics.

\bibitem[{Kaneko et~al.(2022)Kaneko, Takase, Niwa, and Okazaki}]{kaneko2022interpretability_of_gec}
Masahiro Kaneko, Sho Takase, Ayana Niwa, and Naoaki Okazaki. 2022.
\newblock \href {https://doi.org/10.18653/v1/2022.acl-long.496} {Interpretability for language learners using example-based grammatical error correction}.
\newblock In \emph{Proceedings of the 60th Annual Meeting of the Association for Computational Linguistics (Volume 1: Long Papers)}, pages 7176--7187, Dublin, Ireland. Association for Computational Linguistics.

\bibitem[{Katinskaia and Yangarber(2021)}]{katinskaia2021assessing_in_learning}
Anisia Katinskaia and Roman Yangarber. 2021.
\newblock \href {https://aclanthology.org/2021.bea-1.15} {Assessing grammatical correctness in language learning}.
\newblock In \emph{Proceedings of the 16th Workshop on Innovative Use of NLP for Building Educational Applications}, pages 135--146, Online. Association for Computational Linguistics.

\bibitem[{Katsumata and Komachi(2020)}]{katsumata-komachi-2020-bart-basseline-gec}
Satoru Katsumata and Mamoru Komachi. 2020.
\newblock \href {https://aclanthology.org/2020.aacl-main.83} {Stronger baselines for grammatical error correction using a pretrained encoder-decoder model}.
\newblock In \emph{Proceedings of the 1st Conference of the Asia-Pacific Chapter of the Association for Computational Linguistics and the 10th International Joint Conference on Natural Language Processing}, pages 827--832, Suzhou, China. Association for Computational Linguistics.

\bibitem[{Lai et~al.(2022)Lai, Zhou, Zeng, Li, Li, Cao, and Su}]{lai-etal-2022-type-driven-multi-turn}
Shaopeng Lai, Qingyu Zhou, Jiali Zeng, Zhongli Li, Chao Li, Yunbo Cao, and Jinsong Su. 2022.
\newblock \href {https://doi.org/10.18653/v1/2022.findings-acl.254} {Type-driven multi-turn corrections for grammatical error correction}.
\newblock In \emph{Findings of the Association for Computational Linguistics: ACL 2022}, pages 3225--3236, Dublin, Ireland. Association for Computational Linguistics.

\bibitem[{Lewis et~al.(2019)Lewis, Liu, Goyal, Ghazvininejad, Mohamed, Levy, Stoyanov, and Zettlemoyer}]{2019bart}
Mike Lewis, Yinhan Liu, Naman Goyal, Marjan Ghazvininejad, Abdelrahman Mohamed, Omer Levy, Veselin Stoyanov, and Luke Zettlemoyer. 2019.
\newblock \href {http://arxiv.org/abs/1910.13461} {{BART:} denoising sequence-to-sequence pre-training for natural language generation, translation, and comprehension}.
\newblock \emph{CoRR}, abs/1910.13461.

\bibitem[{Li et~al.(2022)Li, Guo, Zhu, Sheng, Jiang, Ren, and Xu}]{li2022seq2action}
Jiquan Li, Junliang Guo, Yongxin Zhu, Xin Sheng, Deqiang Jiang, Bo~Ren, and Linli Xu. 2022.
\newblock \href {https://doi.org/10.1609/aaai.v36i10.21345} {Sequence-to-action: Grammatical error correction with action guided sequence generation}.
\newblock \emph{Proceedings of the AAAI Conference on Artificial Intelligence}, 36(10):10974--10982.

\bibitem[{Li et~al.(2023{\natexlab{a}})Li, Liu, Wang, Gong, Wong, Gao, Huang, and Zhang}]{li-etal-2023-templategec}
Yinghao Li, Xuebo Liu, Shuo Wang, Peiyuan Gong, Derek~F. Wong, Yang Gao, Heyan Huang, and Min Zhang. 2023{\natexlab{a}}.
\newblock \href {https://doi.org/10.18653/v1/2023.acl-long.380} {{T}emplate{GEC}: Improving grammatical error correction with detection template}.
\newblock In \emph{Proceedings of the 61st Annual Meeting of the Association for Computational Linguistics (Volume 1: Long Papers)}, pages 6878--6892, Toronto, Canada. Association for Computational Linguistics.

\bibitem[{Li et~al.(2023{\natexlab{b}})Li, Huang, Ma, Jiang, Li, Zhou, Zheng, and Zhou}]{li2023gpt-chinesegec}
Yinghui Li, Haojing Huang, Shirong Ma, Yong Jiang, Yangning Li, Feng Zhou, Hai-Tao Zheng, and Qingyu Zhou. 2023{\natexlab{b}}.
\newblock On the (in) effectiveness of large language models for chinese text correction.
\newblock \emph{arXiv preprint arXiv:2307.09007}.

\bibitem[{Liao et~al.(2023)Liao, Eskimez, Lu, Shi, Gong, Shou, Qu, and Zeng}]{liao2023improving_asr_by_gec}
Junwei Liao, Sefik Eskimez, Liyang Lu, Yu~Shi, Ming Gong, Linjun Shou, Hong Qu, and Michael Zeng. 2023.
\newblock \href {https://doi.org/10.1145/3557894} {Improving readability for automatic speech recognition transcription}.
\newblock \emph{ACM Trans. Asian Low-Resour. Lang. Inf. Process.}, 22(5).

\bibitem[{Lichtarge et~al.(2020)Lichtarge, Alberti, and Kumar}]{lichtarge2020dataweightedtraining_gec}
Jared Lichtarge, Chris Alberti, and Shankar Kumar. 2020.
\newblock \href {https://doi.org/10.1162/tacl_a_00336} {Data weighted training strategies for grammatical error correction}.
\newblock \emph{Transactions of the Association for Computational Linguistics}, 8:634--646.

\bibitem[{Lin et~al.(2020)Lin, Goyal, Girshick, He, and Dollár}]{lin2017focalloss}
Tsung-Yi Lin, Priya Goyal, Ross Girshick, Kaiming He, and Piotr Dollár. 2020.
\newblock \href {https://doi.org/10.1109/TPAMI.2018.2858826} {Focal loss for dense object detection}.
\newblock \emph{IEEE Transactions on Pattern Analysis and Machine Intelligence}, 42(2):318--327.

\bibitem[{Loem et~al.(2023)Loem, Kaneko, Takase, and Okazaki}]{loem2023gpt3-gec-en}
Mengsay Loem, Masahiro Kaneko, Sho Takase, and Naoaki Okazaki. 2023.
\newblock Exploring effectiveness of gpt-3 in grammatical error correction: A study on performance and controllability in prompt-based methods.
\newblock \emph{arXiv preprint arXiv:2305.18156}.

\bibitem[{Mallinson et~al.(2020)Mallinson, Severyn, Malmi, and Garrido}]{mallinson-etal-2020-felix-text-editing}
Jonathan Mallinson, Aliaksei Severyn, Eric Malmi, and Guillermo Garrido. 2020.
\newblock \href {https://doi.org/10.18653/v1/2020.findings-emnlp.111} {{FELIX}: Flexible text editing through tagging and insertion}.
\newblock In \emph{Findings of the Association for Computational Linguistics: EMNLP 2020}, pages 1244--1255, Online. Association for Computational Linguistics.

\bibitem[{Malmi et~al.(2019)Malmi, Krause, Rothe, Mirylenka, and Severyn}]{malmi-etal-2019-lasertagger}
Eric Malmi, Sebastian Krause, Sascha Rothe, Daniil Mirylenka, and Aliaksei Severyn. 2019.
\newblock \href {https://doi.org/10.18653/v1/D19-1510} {Encode, tag, realize: High-precision text editing}.
\newblock In \emph{Proceedings of the 2019 Conference on Empirical Methods in Natural Language Processing and the 9th International Joint Conference on Natural Language Processing (EMNLP-IJCNLP)}, pages 5054--5065, Hong Kong, China. Association for Computational Linguistics.

\bibitem[{Mita et~al.(2020)Mita, Kiyono, Kaneko, Suzuki, and Inui}]{mita-etal-2020-noise-refinement}
Masato Mita, Shun Kiyono, Masahiro Kaneko, Jun Suzuki, and Kentaro Inui. 2020.
\newblock \href {https://doi.org/10.18653/v1/2020.findings-emnlp.26} {A self-refinement strategy for noise reduction in grammatical error correction}.
\newblock In \emph{Findings of the Association for Computational Linguistics: EMNLP 2020}, pages 267--280, Online. Association for Computational Linguistics.

\bibitem[{Ng et~al.(2014)Ng, Wu, Briscoe, Hadiwinoto, Susanto, and Bryant}]{ng-etal-2014-conll14}
Hwee~Tou Ng, Siew~Mei Wu, Ted Briscoe, Christian Hadiwinoto, Raymond~Hendy Susanto, and Christopher Bryant. 2014.
\newblock \href {https://doi.org/10.3115/v1/W14-1701} {The {C}o{NLL}-2014 shared task on grammatical error correction}.
\newblock In \emph{Proceedings of the Eighteenth Conference on Computational Natural Language Learning: Shared Task}, pages 1--14, Baltimore, Maryland. Association for Computational Linguistics.

\bibitem[{Omelianchuk et~al.(2020)Omelianchuk, Atrasevych, Chernodub, and Skurzhanskyi}]{omelianchuk2020gector_gec}
Kostiantyn Omelianchuk, Vitaliy Atrasevych, Artem Chernodub, and Oleksandr Skurzhanskyi. 2020.
\newblock \href {https://doi.org/10.18653/v1/2020.bea-1.16} {{GECT}o{R} {--} grammatical error correction: Tag, not rewrite}.
\newblock In \emph{Proceedings of the Fifteenth Workshop on Innovative Use of NLP for Building Educational Applications}, pages 163--170, Seattle, WA, USA → Online. Association for Computational Linguistics.

\bibitem[{Ouyang et~al.(2022)Ouyang, Wu, Jiang, Almeida, Wainwright, Mishkin, Zhang, Agarwal, Slama, Ray, Schulman, Hilton, Kelton, Miller, Simens, Askell, Welinder, Christiano, Leike, and Lowe}]{ouyang2022GPT3_5_RLHF}
Long Ouyang, Jeffrey Wu, Xu~Jiang, Diogo Almeida, Carroll Wainwright, Pamela Mishkin, Chong Zhang, Sandhini Agarwal, Katarina Slama, Alex Ray, John Schulman, Jacob Hilton, Fraser Kelton, Luke Miller, Maddie Simens, Amanda Askell, Peter Welinder, Paul~F Christiano, Jan Leike, and Ryan Lowe. 2022.
\newblock \href {https://proceedings.neurips.cc/paper_files/paper/2022/file/b1efde53be364a73914f58805a001731-Paper-Conference.pdf} {Training language models to follow instructions with human feedback}.
\newblock In \emph{Advances in Neural Information Processing Systems}, volume~35, pages 27730--27744. Curran Associates, Inc.

\bibitem[{Qu and Wu(2023)}]{qu2023llm-chinesegec}
Fanyi Qu and Yunfang Wu. 2023.
\newblock Evaluating the capability of large-scale language models on chinese grammatical error correction task.
\newblock \emph{arXiv preprint arXiv:2307.03972}.

\bibitem[{Raffel et~al.(2020)Raffel, Shazeer, Roberts, Lee, Narang, Matena, Zhou, Li, and Liu}]{2020t5}
Colin Raffel, Noam Shazeer, Adam Roberts, Katherine Lee, Sharan Narang, Michael Matena, Yanqi Zhou, Wei Li, and Peter~J. Liu. 2020.
\newblock \href {http://jmlr.org/papers/v21/20-074.html} {Exploring the limits of transfer learning with a unified text-to-text transformer}.
\newblock \emph{Journal of Machine Learning Research}, 21(140):1--67.

\bibitem[{Rei and Yannakoudakis(2016)}]{rei2016compositional_detection_gec}
Marek Rei and Helen Yannakoudakis. 2016.
\newblock \href {https://doi.org/10.18653/v1/P16-1112} {Compositional sequence labeling models for error detection in learner writing}.
\newblock In \emph{Proceedings of the 54th Annual Meeting of the Association for Computational Linguistics (Volume 1: Long Papers)}, pages 1181--1191, Berlin, Germany. Association for Computational Linguistics.

\bibitem[{Rothe et~al.(2021)Rothe, Mallinson, Malmi, Krause, and Severyn}]{rothe2021clang8_gec}
Sascha Rothe, Jonathan Mallinson, Eric Malmi, Sebastian Krause, and Aliaksei Severyn. 2021.
\newblock \href {https://doi.org/10.18653/v1/2021.acl-short.89} {A simple recipe for multilingual grammatical error correction}.
\newblock In \emph{Proceedings of the 59th Annual Meeting of the Association for Computational Linguistics and the 11th International Joint Conference on Natural Language Processing (Volume 2: Short Papers)}, pages 702--707, Online. Association for Computational Linguistics.

\bibitem[{Stahlberg and Kumar(2020)}]{stahlberg2020seq2edits_gec}
Felix Stahlberg and Shankar Kumar. 2020.
\newblock \href {https://doi.org/10.18653/v1/2020.emnlp-main.418} {{S}eq2{E}dits: Sequence transduction using span-level edit operations}.
\newblock In \emph{Proceedings of the 2020 Conference on Empirical Methods in Natural Language Processing (EMNLP)}, pages 5147--5159, Online. Association for Computational Linguistics.

\bibitem[{Stahlberg and Kumar(2021)}]{stahlberg-kumar-2021-synthetic-tag-corrupted}
Felix Stahlberg and Shankar Kumar. 2021.
\newblock \href {https://aclanthology.org/2021.bea-1.4} {Synthetic data generation for grammatical error correction with tagged corruption models}.
\newblock In \emph{Proceedings of the 16th Workshop on Innovative Use of NLP for Building Educational Applications}, pages 37--47, Online. Association for Computational Linguistics.

\bibitem[{Sun et~al.(2023)Sun, Gao, Wu, Fang, Cao, and Du}]{sun2023htec_text_data_by_gec}
Hanbo Sun, Jian Gao, Xiaomin Wu, Anjie Fang, Cheng Cao, and Zheng Du. 2023.
\newblock Htec: Human transcription error correction.
\newblock \emph{arXiv preprint arXiv:2309.10089}.

\bibitem[{Sun et~al.(2021)Sun, Ge, Wei, and Wang}]{sun-etal-2021-instantaneous-gec}
Xin Sun, Tao Ge, Furu Wei, and Houfeng Wang. 2021.
\newblock \href {https://doi.org/10.18653/v1/2021.acl-long.462} {Instantaneous grammatical error correction with shallow aggressive decoding}.
\newblock In \emph{Proceedings of the 59th Annual Meeting of the Association for Computational Linguistics and the 11th International Joint Conference on Natural Language Processing (Volume 1: Long Papers)}, pages 5937--5947, Online. Association for Computational Linguistics.

\bibitem[{Touvron et~al.(2023)Touvron, Martin, Stone, Albert, Almahairi, Babaei, Bashlykov, Batra, Bhargava, Bhosale et~al.}]{touvron2023llama2}
Hugo Touvron, Louis Martin, Kevin Stone, Peter Albert, Amjad Almahairi, Yasmine Babaei, Nikolay Bashlykov, Soumya Batra, Prajjwal Bhargava, Shruti Bhosale, et~al. 2023.
\newblock Llama 2: Open foundation and fine-tuned chat models.
\newblock \emph{arXiv preprint arXiv:2307.09288}.

\bibitem[{Wang et~al.(2021)Wang, Wang, Dang, Liu, and Liu}]{wang2021comprehensive_survey_gec}
Yu~Wang, Yuelin Wang, Kai Dang, Jie Liu, and Zhuo Liu. 2021.
\newblock \href {https://doi.org/10.1145/3474840} {A comprehensive survey of grammatical error correction}.
\newblock \emph{ACM Trans. Intell. Syst. Technol.}, 12(5).

\bibitem[{Xu et~al.(2022)Xu, Wu, Peng, Fu, and Cai}]{xu-etal-2022-fcgec}
Lvxiaowei Xu, Jianwang Wu, Jiawei Peng, Jiayu Fu, and Ming Cai. 2022.
\newblock \href {https://doi.org/10.18653/v1/2022.findings-emnlp.137} {{FCGEC}: Fine-grained corpus for {C}hinese grammatical error correction}.
\newblock In \emph{Findings of the Association for Computational Linguistics: EMNLP 2022}, pages 1900--1918, Abu Dhabi, United Arab Emirates. Association for Computational Linguistics.

\bibitem[{Yakovlev et~al.(2023)Yakovlev, Podolskiy, Bout, Nikolenko, and Piontkovskaya}]{yakovlev-etal-2023-gec-depend}
Konstantin Yakovlev, Alexander Podolskiy, Andrey Bout, Sergey Nikolenko, and Irina Piontkovskaya. 2023.
\newblock \href {https://doi.org/10.18653/v1/2023.acl-long.86} {{GEC}-{D}e{P}en{D}: Non-autoregressive grammatical error correction with decoupled permutation and decoding}.
\newblock In \emph{Proceedings of the 61st Annual Meeting of the Association for Computational Linguistics (Volume 1: Long Papers)}, pages 1546--1558, Toronto, Canada. Association for Computational Linguistics.

\bibitem[{Yang et~al.(2023)Yang, Xiao, Wang, Zhang, Bian, Yin, Lv, Pan, Wang, Yan et~al.}]{yang2023baichuan2}
Aiyuan Yang, Bin Xiao, Bingning Wang, Borong Zhang, Ce~Bian, Chao Yin, Chenxu Lv, Da~Pan, Dian Wang, Dong Yan, et~al. 2023.
\newblock Baichuan 2: Open large-scale language models.
\newblock \emph{arXiv preprint arXiv:2309.10305}.

\bibitem[{Yannakoudakis et~al.(2011)Yannakoudakis, Briscoe, and Medlock}]{yannakoudakis-etal-2011-fce}
Helen Yannakoudakis, Ted Briscoe, and Ben Medlock. 2011.
\newblock \href {https://aclanthology.org/P11-1019} {A new dataset and method for automatically grading {ESOL} texts}.
\newblock In \emph{Proceedings of the 49th Annual Meeting of the Association for Computational Linguistics: Human Language Technologies}, pages 180--189, Portland, Oregon, USA. Association for Computational Linguistics.

\bibitem[{Yuan et~al.(2021{\natexlab{a}})Yuan, Taslimipoor, Davis, and Bryant}]{yuan-etal-2021-two_systems_detection_correction_gec}
Zheng Yuan, Shiva Taslimipoor, Christopher Davis, and Christopher Bryant. 2021{\natexlab{a}}.
\newblock \href {https://doi.org/10.18653/v1/2021.emnlp-main.687} {{M}ulti-class grammatical error detection for correction: {A} tale of two systems}.
\newblock In \emph{Proceedings of the 2021 Conference on Empirical Methods in Natural Language Processing}, pages 8722--8736, Online and Punta Cana, Dominican Republic. Association for Computational Linguistics.

\bibitem[{Yuan et~al.(2021{\natexlab{b}})Yuan, Taslimipoor, Davis, and Bryant}]{yuan2021multiclass_detection_correction}
Zheng Yuan, Shiva Taslimipoor, Christopher Davis, and Christopher Bryant. 2021{\natexlab{b}}.
\newblock \href {https://doi.org/10.18653/v1/2021.emnlp-main.687} {{M}ulti-class grammatical error detection for correction: {A} tale of two systems}.
\newblock In \emph{Proceedings of the 2021 Conference on Empirical Methods in Natural Language Processing}, pages 8722--8736, Online and Punta Cana, Dominican Republic. Association for Computational Linguistics.

\bibitem[{Zeng et~al.(2022)Zeng, Liu, Du, Wang, Lai, Ding, Yang, Xu, Zheng, Xia et~al.}]{zeng2022glm_130b}
Aohan Zeng, Xiao Liu, Zhengxiao Du, Zihan Wang, Hanyu Lai, Ming Ding, Zhuoyi Yang, Yifan Xu, Wendi Zheng, Xiao Xia, et~al. 2022.
\newblock Glm-130b: An open bilingual pre-trained model.
\newblock \emph{arXiv preprint arXiv:2210.02414}.

\bibitem[{Zhang et~al.(2023)Zhang, Kamigaito, and Okumura}]{zhang-etal-2023-transformer-reranker}
Ying Zhang, Hidetaka Kamigaito, and Manabu Okumura. 2023.
\newblock \href {https://doi.org/10.18653/v1/2023.findings-acl.234} {Bidirectional transformer reranker for grammatical error correction}.
\newblock In \emph{Findings of the Association for Computational Linguistics: ACL 2023}, pages 3801--3825, Toronto, Canada. Association for Computational Linguistics.

\bibitem[{Zhang et~al.(2022{\natexlab{a}})Zhang, Li, Bao, Li, Zhang, Li, Huang, and Zhang}]{zhang-etal-2022-mucgec}
Yue Zhang, Zhenghua Li, Zuyi Bao, Jiacheng Li, Bo~Zhang, Chen Li, Fei Huang, and Min Zhang. 2022{\natexlab{a}}.
\newblock \href {https://doi.org/10.18653/v1/2022.naacl-main.227} {{M}u{CGEC}: a multi-reference multi-source evaluation dataset for {C}hinese grammatical error correction}.
\newblock In \emph{Proceedings of the 2022 Conference of the North American Chapter of the Association for Computational Linguistics: Human Language Technologies}, pages 3118--3130, Seattle, United States. Association for Computational Linguistics.

\bibitem[{Zhang et~al.(2022{\natexlab{b}})Zhang, Zhang, Li, Bao, Li, and Zhang}]{zhang-etal-2022-syngec}
Yue Zhang, Bo~Zhang, Zhenghua Li, Zuyi Bao, Chen Li, and Min Zhang. 2022{\natexlab{b}}.
\newblock \href {https://doi.org/10.18653/v1/2022.emnlp-main.162} {{S}yn{GEC}: Syntax-enhanced grammatical error correction with a tailored {GEC}-oriented parser}.
\newblock In \emph{Proceedings of the 2022 Conference on Empirical Methods in Natural Language Processing}, pages 2518--2531, Abu Dhabi, United Arab Emirates. Association for Computational Linguistics.

\bibitem[{Zhao et~al.(2019)Zhao, Wang, Shen, Jia, and Liu}]{zhao-etal-2019-seq2seq-copy-augmented}
Wei Zhao, Liang Wang, Kewei Shen, Ruoyu Jia, and Jingming Liu. 2019.
\newblock \href {https://doi.org/10.18653/v1/N19-1014} {Improving grammatical error correction via pre-training a copy-augmented architecture with unlabeled data}.
\newblock In \emph{Proceedings of the 2019 Conference of the North {A}merican Chapter of the Association for Computational Linguistics: Human Language Technologies, Volume 1 (Long and Short Papers)}, pages 156--165, Minneapolis, Minnesota. Association for Computational Linguistics.

\bibitem[{Zhao et~al.(2018)Zhao, Jiang, Sun, and Wan}]{zhao2018nlpcc-gec}
Yuanyuan Zhao, Nan Jiang, Weiwei Sun, and Xiaojun Wan. 2018.
\newblock Overview of the nlpcc 2018 shared task: Grammatical error correction.
\newblock In \emph{Natural Language Processing and Chinese Computing}, pages 439--445, Cham. Springer International Publishing.

\bibitem[{Zhou et~al.(2023)Zhou, Liu, Li, Zhang, Zhang, Li, Zhang, and Huang}]{zhou-etal-2023-improving-seq2seq-decoding-interventions}
Houquan Zhou, Yumeng Liu, Zhenghua Li, Min Zhang, Bo~Zhang, Chen Li, Ji~Zhang, and Fei Huang. 2023.
\newblock \href {https://doi.org/10.18653/v1/2023.findings-emnlp.495} {Improving {S}eq2{S}eq grammatical error correction via decoding interventions}.
\newblock In \emph{Findings of the Association for Computational Linguistics: EMNLP 2023}, pages 7393--7405, Singapore. Association for Computational Linguistics.

\end{thebibliography}

\clearpage
\appendix

\section{Dataset}
\label{sec:appendix-dataset}
\subsection{Dataset Statistics}

\begin{table}[ht]
\centering
\resizebox{\columnwidth}{!}{
\begin{tabular}{llll}
\hline
\textbf{Dataset}                & \textbf{\#Sentences} & \textbf{Usage} & \textbf{As training data of} \\ \hline
\textbf{C4-200M}                & 183,894,319          & Pretraining    & DeCoGLM, DeGLM, CoGLM        \\
\textbf{Synthetic-CH}           & 33,166,047           & Pretraining    & DeCoGLM, DeGLM, CoGLM        \\ \hline
\textbf{CLang8(EN)}              & 2,372,119            & Fine-tuning    & DeCoGLM, DeGLM, CoGLM        \\
\textbf{FCE} \textit{all}       & 33,236               & Fine-tuning    & CoGLM (10B)                  \\
\textbf{NUCLE}                  & 57,157               & Fine-tuning    & CoGLM (10B)                  \\
\textbf{W\&I+LOCNESS}           & 34,308               & Fine-tuning    & CoGLM (10B)                  \\
\textbf{Lang8 (CH)}               & 1,092,285            & Fine-tuning    & All                          \\
\textbf{HSK}                    & 95,320               & Fine-tuning    & All                          \\
\textbf{FCGEC} \textit{train}   & 36,341               & Fine-tuning    & CoGLM (10B)                  \\ \hline
\textbf{BEA19} \textit{dev}     & 4,384                & Validation     & -                            \\
\textbf{MuCGEC} \textit{dev}    & 1,137                & Validation     & -                            \\
\textbf{FCGEC} \textit{dev}     & 2,000                & Validation     & -                            \\ \hline
\textbf{CoNLL-14} \textit{test} & 1,312                & Testing        & -                            \\
\textbf{BEA19} \textit{test}    & 4,477                & Testing        & -                            \\
\textbf{MuCGEC} \textit{test}   & 6,000                & Testing        & -                            \\
\textbf{FCGEC} \textit{test}    & 3,000                & Testing        & -                            \\ \hline
\end{tabular}
}

\caption{
Dataset statistics. The rightmost column indicates the models that utilize the respective dataset; "All" signifies that DeCoGLM, DeGLM, CoGLM, and CoGLM (10B) all used the dataset as the training set.
}
\label{tab:data-statistics}
\end{table}

In the experiments described in Section \ref{sec:data-and-evaluation}, the datasets used are outlined in Table \ref{tab:data-statistics}. Due to constraints on our computational resources, the CoGLM (10B) models are fine-tuned on relatively smaller datasets, and the models are not pre-trained on synthetic datasets.

\subsection{Dataset for Training}

As shown in Table \ref{tab:data-statistics}, for relatively small models, we first pretrain using the publicly available C4-200M English GEC synthetic dataset and our synthesized Chinese GEC dataset to obtain two pretrained models. Subsequently, the English model is fine-tuned using the CLang8 dataset, while the Chinese model is fine-tuned using the Lang8 Chinese dataset and the FCGEC dataset to yield two individual models. For the results of the models used for comparison in the primary results of Table \ref{main-results}, the GECToR and BART results on the two Chinese datasets are reproduced according to our training procedure, while the rest are as reported in the original papers, where they utilized various training data configurations. Most models are fine-tuned using the NUCLE, W\&I+LOCNESS, and FCE datasets. Besides, GECToR uses the PIE-9M as the pretraining dataset \citep{awasthi2019parallel_edit_seq2edit_gec}.

For LLMs, we did not perform pretraining; instead, we directly applied the LoRA method using the NUCLE, W\&I+LOCNESS, and FCE datasets for the English model. For Chinese, we trained two models using Lang8 (CH) and FCGEC. The training data for CoGLM (10B) and other LLMs used for comparison are completely consistent.

\begin{table*}[ht]
\centering
\resizebox{\textwidth}{!}{
\begin{tabular}{c|l|l|l}
\hline
\textbf{Stage}                 & \textbf{Items}                          & \textbf{Example 1}                                              & \textbf{Example 2}                                   \\ \hline
\multirow{6}{*}{\textbf{SFT1}} & Source Text $\boldsymbol x_s$           & <s>The every male employees were standing in the back row .</s> & <s>They are covered with rust so bad .</s>           \\ \cline{2-4} 
                               & Target Text $\boldsymbol y$             & <s>All the male employees were standing in the back row .</s>   & <s>They are covered with rust so badly .</s>         \\ \cline{2-4} 
                               & Masked Text $\boldsymbol x_m$           & <s>[MASK] male employees were standing in the back row .</s>    & <s>They are covered with rust so [MASK] .</s>        \\ \cline{2-4} 
                               & Text Pieces Input                       & <|startofpiece|> All the                                        & <|startofpiece|> badly                               \\ \cline{2-4} 
                               & Text Pieces Target                      & All the <|endofpiece|>                                          & badly <|endofpiece|>                                 \\ \cline{2-4} 
                               & Detection Labels                        & K E E K K K K K K K K K K                                       & K K K K K K K E K K                                  \\ \hline
\multirow{5}{*}{\textbf{SFT2}} & Detections by SFT1                      & K E E K E E K K K K K K K                                       & K K K K K I K K K K                                  \\ \cline{2-4} 
                               & Merged Detecions                        & K E E K E E K K K K K K K                                       & K K K K K I K E K K                                  \\ \cline{2-4} 
                               & Masked Text $\boldsymbol x^{\prime}_m $ & <s>[MASK] male [MASK] standing in the back row.</s>             & <s>They are covered with rust [MASK] so [MASK] .</s> \\ \cline{2-4} 
                               & Text Pieces Input                       & <|startofpiece|> All the <|startofpiece|>  employees were       & <|startofpiece|> <|startofpiece|> badly              \\ \cline{2-4} 
                               & Text Pieces Target                      & All the <|endofpiece|>  employees were <|endofpiece|>           & <|endofpiece|> badly <|endofpiece|>                  \\ \hline
\end{tabular}
}
\caption{
Examples of training data from CLang8 dataset in two fine-tuning stages. In detection labels, K=KEEP, E=ERROR and I=INSERT.
}
\label{tab:data-examples}
\end{table*}
\begin{table*}[ht]
\centering
\resizebox{0.9\textwidth}{!}{
\begin{tabular}{lcccc}
\hline
\textbf{Configuration}           & \textbf{EN Pretrain}       & \textbf{EN finetune}       & \textbf{CH Pretrain} & \textbf{CH finetune}                        \\ \hline
\multicolumn{5}{c}{\textbf{DeCoGLM-Training}}                                                                                                                   \\ \hline
\textbf{Backbone}                & \multicolumn{2}{c}{GLM-RoBERTa-large \citep{du2021glm}} & \multicolumn{2}{c}{GLM-large-chinese \citep{du2021glm}}            \\
\textbf{Backbone Parameters}     & \multicolumn{2}{c}{335M}                                & \multicolumn{2}{c}{335M}                                           \\
\textbf{Batch size}              & 12                         & 12                         & 12                   & 12                                          \\
\textbf{Update frequecy}         & 10                         & 20                         & 8                    & 8(M), 10(F)                                 \\
\textbf{Max epochs}              & (20M iterations)             & 20                         & 2                    & 10(M), 20(F)                                \\
\textbf{Evaluation key (SFT1)}   & -                          & AD-Accuracy                & AD-Accuracy          & AD-Accuracy                                 \\
\textbf{Evaluation key (SFT2)}   & -                          & General-Accuracy           & -                    & General-Accuracy                            \\
\textbf{Evaluation interval}     & 10000                      & 2000                       & 4000                 & 2000(M), 200(F)                             \\
\textbf{Early stop}              & -                          & 10                         & -                    & 10                                          \\
\textbf{Max source text length}  & 128                        & 128                        & 128                  & 128                                         \\
\textbf{Warm-up steps (SFT1)}    & 10000                      & 1000                       & 1000                 & 1000(M), 200(F)                             \\
\textbf{Warm-up steps (SFT2)}    & -                          & 1000                       & -                    & 1000(M), 200(F)                             \\
\textbf{Weight Decay}            & $1\times 10^{-4}$          & $1\times 10^{-4}$          & $1\times 10^{-4}$    & $1\times 10^{-4}$                           \\
\textbf{Learning rate scheduler} & Polynomial                 & Polynomial                 & Polynomial           & Polynomial                                  \\
\textbf{Learning rate (SFT1)}    & $2\times 10^{-5}$          & $3\times 10^{-6}$          & $2\times 10^{-5}$    & $1\times 10^{-5}$ (M), $4\times 10^{-5}$(F) \\
\textbf{Learning rate (SFT2)}    & -                          & $1\times 10^{-6}$          & -                    & $5\times 10^{-6}$ (M), $1\times 10^{-5}$(F) \\ \hline
\multicolumn{5}{c}{\textbf{DeCoGLM-Inference}}                                                                                                                  \\ \hline
\textbf{KEEP threshold}          & \multicolumn{2}{c}{0.38}                                & \multicolumn{2}{c}{None}                                           \\
\textbf{ERROR lower bound}       & \multicolumn{2}{c}{0.5}                                 & \multicolumn{2}{c}{None}                                           \\
\textbf{INSERT lower bound}      & \multicolumn{2}{c}{0.6}                                 & \multicolumn{2}{c}{None}                                           \\
\textbf{Beam size}               & \multicolumn{2}{c}{3}                                   & \multicolumn{2}{c}{3}                                              \\
\textbf{Max tokens per piece}    & \multicolumn{2}{c}{10}                                  & \multicolumn{2}{c}{10}                                             \\ \hline
\end{tabular}
}
\caption{\label{tab:train-settings}
The model hyper-parameters of proposed DeCoGLM. Both pretraining and fine-tuning configurations are presented. EN and CH represent English models and Chinese models, respectively. In the settings of Chinese fine-tuned models, M and F represent models for MuCGEC and FCGEC, respectively. The bottom of the table presents the hyper-parameters of inference.
}
\end{table*}

\subsection{Dataset Examples}
In Sections \ref{sec:error-detection} and \ref{sec:local-error-correction}, we describe the construction of training data. By aligning the source text with the target text, we derive error detection labels and masked text, thereby constructing training samples as illustrated in Figure \ref{fig:model}. In Section \ref{sec:multi-task-organization}, we elaborate on a two-stage supervised fine-tuning approach, where the training data for the second stage is reconstructed based on the detection predictions made by the model trained in the first stage. During data construction, model-induced false positives for ERROR and INSERT are incorporated to generate new masked text and corresponding text pieces. It is crucial to note that this process is solely aimed at creating new masked text to enhance the model's ability to address false positives during the correction phase, while the detection labels used in training remain unchanged. Examples of the constructed training data are provided in Table \ref{tab:data-examples}, where "<s>" denotes the "begin of sentence" token and "</s>" represents the "end of sentence" token. For the sake of brevity, these tokens are omitted in the content of this paper except in Figure \ref{fig:model}.

\begin{table*}
\centering
\resizebox{\textwidth}{!}{
\begin{tabular}{l|l}
\hline
\textbf{Mode}     & \textbf{Prompt}                                                                                                                                                                                                                                                                                                                                                                                                                                                                                                                                                                                                                                       \\ \hline
\textbf{ZeroShot} & \begin{tabular}[c]{@{}l@{}}Reply with a corrected version of the input sentence with all grammatical and spelling errors fixed. If there are no errors, \\ reply with a copy of the original sentence.\\ \\ Input sentence: [TEXT]\\ Corrected sentence:\end{tabular}                                                                                                                                                                                                                                                                                                                                                                                 \\ \hline
\textbf{+DeGLM}   & \begin{tabular}[c]{@{}l@{}}Reply with a corrected version of the input sentence with all grammatical and spelling errors fixed. If there are no errors, \\ reply with a copy of the original sentence.\\ Hint: We have detected some possible grammatical errors and replaced every error span with a [MASK] to get a masked \\ sentence, you can reference the masked sentence to give final corrected sentence. If there is no [MASK] in the masked sentence, \\ it means that we have not detected any grammatical errors in the input sentence. \\ \\ Input sentence: [TEXT]\\ Masked Sentence: [MASKED\_TEXT]\\ Corrected sentence:\end{tabular} \\ \hline
\end{tabular}
}
\caption{
GPT-4 prompts used in experiments, following \citet{coyne2023gpt-gec-en}.
}
\label{tab:prompt}
\end{table*}
\begin{table*}
\centering
\resizebox{0.7\linewidth}{!}{
\begin{tabular}{cc|cc|ccc}
\hline
\textbf{}         & \textbf{}          & \multicolumn{2}{c|}{$\mathbf{F_{0.5}}$ \textbf{on test set}} & \multicolumn{3}{c}{\textbf{Average inference time per sample (ms)}} \\
\textbf{Backbone} & \textbf{Structure} & \textbf{CoNLL-14}              & \textbf{BEA-19}             & \textbf{Detection}     & \textbf{Correction}    & \textbf{Total}    \\ \hline
GLM-Roberta       & De-Co              & \textbf{64.71}                 & \textbf{70.85}              & \textbf{14.5}          & 69.1                   & 83.6              \\
BART-large        & De-Co              & 62.24                          & 67.97                       & 17.1                   & \textbf{43.4}          & \textbf{60.5}     \\
BART-large        & Seq2Seq            & 64.46                          & 67.94                       & \textbf{-}             & 266.2                  & 266.2             \\ \hline
\end{tabular}
}
\caption{
Time consumed in inference. De-Co represents the proposed detection-correction structure.
}
\label{tab:time}
\end{table*}


\section{Details of Experiments}

\subsection{Loss Weight}
\label{sec:appendix-weights}

\begin{figure}[h]
    \centering
    \includegraphics[width=0.9\columnwidth]{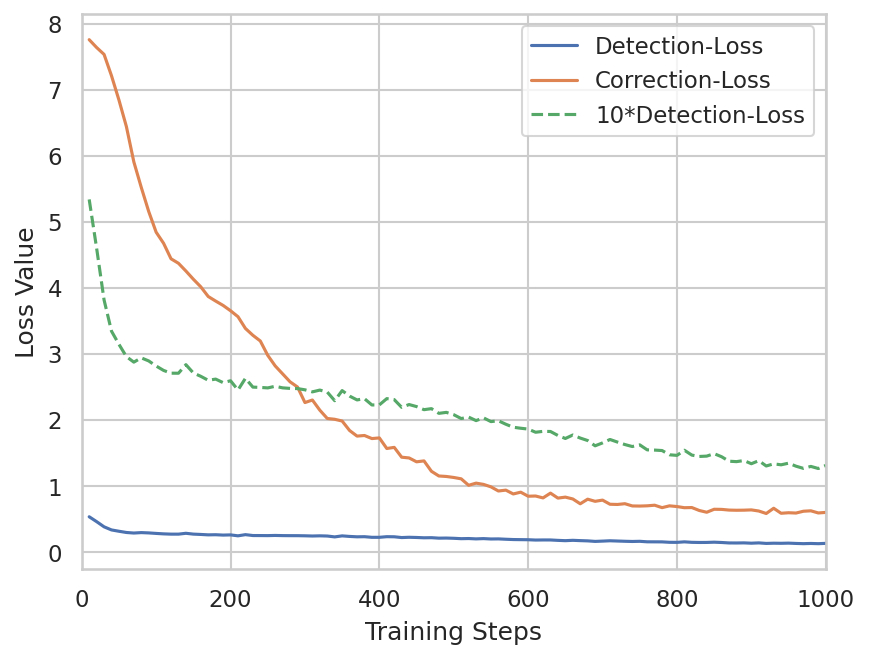}
    \caption{Loss curves in standard training condition.}
    \label{fig:loss-curve}
\end{figure}

We pre-determine the weights in multi-task learning by intuitively observing the scales of two losses. This preliminary experiment was conducted on the CLang8 dataset, and the loss curves are depicted in Figure \ref{fig:loss-curve}. It is evident from the figure that the detection loss $\ell_D$ and correction loss $\ell_C$ differ by roughly an order of magnitude. Consequently, we initially set $w_D=10$, determine the weights for ERROR and INSERT categories in Focal Loss denoted by \(\alpha_{EI}\), and subsequently test whether $w_D=10$ is an optimal choice, as discussed in Section \ref{sec:weight-study}. 

\subsection{Model Configurations}
\label{sec:appendix-train-settings}

The training configurations for the integrated detection-correction model (DeCoGLM) and the parameters used during inference are presented in Table \ref{tab:train-settings}. To conserve computational resources during training, early stopping is employed, which requires the pre-definition of evaluation metrics on the validation set. Two primary metrics are utilized: (1) AD-Accuracy, defined as the sum of the recall for ERROR and INSERT and the accuracy of next token prediction by GLM, aiming to reinforce the aggressive detection principle mentioned in Section \ref{sec:error-detection}; (2) General-Accuracy, the geometric mean between the recall for the three detection labels and the accuracy of next token prediction by GLM. The configurations for training the separate models, DeGLM and CoGLM, are similar to those in Table \ref{tab:train-settings}. The pre-trained models include \texttt{glm-roberta-large}, \texttt{glm-large-chinese}, \texttt{glm-10b}, and \texttt{glm-10b-chinese}, accessible through HuggingFace\footnote{\href{https://huggingface.co}{https://huggingface.co}}. We implement all the designed models using PyTorch, including DeCoGLM, DeGLM, and CoGLM. 

All models are trained by the Trainer from the transformers\footnote{\href{https://huggingface.co/docs/transformers/index}{https://huggingface.co/docs/transformers/index}} package in Python, on NVIDIA RTX 4090 GPUs. Due to resource constraints, all experiments are conducted with a fixed random seed (111), and single-run results are reported. We adopt the approach recommended by \citet{rothe2021clang8_gec} to post-process the model's predictions on English test datasets, aiming to ensure greater alignment of tokenization with the evaluation data.

\subsection{GPT-4 Prompts}

The prompts utilized during the inference of GPT-4 are illustrated in Table \ref{tab:prompt}. For the Chinese tasks, the prompts are the direct translation of the corresponding English prompts. The API version of GPT-4 used in this paper is Preview-0315.

\section{Inference Speed}
\label{sec:inference-speed}

We conduct a brief evaluation of the inference speed of our proposed detection-correction structure, and the average inference speeds on the CoNLL-14 and BEA-19 test sets are presented in Table \ref{tab:time}. The models are trained exclusively on the CLang8 dataset, and during the inference phase, no hyperparameters are adjusted, utilizing only beam search. Our proposed model achieves slightly better performance while maintaining a faster inference speed ($\approx$3x) than the Seq2Seq model. The experiments are conducted on an NVIDIA RTX 4090 GPU, with the same constrained batch size of 1 during inference.

\end{document}